\documentclass[10pt]{article}

\usepackage[accepted]{tmlr}

\usepackage{amsmath,amsfonts,bm}

\def\eqref#1{equation~\ref{#1}}

\def\1{\bm{1}}

\DeclareMathAlphabet{\mathsfit}{\encodingdefault}{\sfdefault}{m}{sl}
\SetMathAlphabet{\mathsfit}{bold}{\encodingdefault}{\sfdefault}{bx}{n}

\usepackage[T1]{fontenc}
\usepackage[utf8]{inputenc}
\usepackage{xspace}
\usepackage{listings}
\usepackage{mdframed}
\usepackage{subcaption}
\usepackage{multirow}
\usepackage{pifont}
\usepackage{booktabs}
\usepackage{arydshln}
\usepackage{algorithm}
\usepackage{algpseudocode}
\usepackage{algpseudocodex}
\usepackage{etoolbox}
\usepackage[most]{tcolorbox}
\usepackage[dvipsnames,table]{xcolor}
\usepackage[colorlinks, allcolors=black]{hyperref}

\newcommand{\pefuse}{\textsc{PeFuse}\xspace}

\newcommand{\cbox}[2][]{
\begin{tcolorbox}[
    colback=white, 
    colframe=blue!15!white,
    halign=left,
    valign=center, 
    center title,
    coltitle=black,
    fonttitle=\bfseries,
    title={#1}]
#2
\end{tcolorbox}
}

\algrenewcommand\algorithmicrequire{\textbf{Input:}}
\algrenewcommand\algorithmicensure{\textbf{Output:}}
\algnewcommand{\Requirements}{\item[\textbf{Requirements:}]}

\title{Training-Free Pseudo-Fusion for Composed Image Retrieval with Diffusion Models and Multimodal Large Language Models}

\author{\name Fan Xu \email fan.xu@uni.lu \\
  \addr Department of Computer Science \\
  University of Luxembourg \\
  \AND 
  \name Luis A. Leiva \email luis.leiva@uni.lu \\
  \addr Department of Computer Science \\
  University of Luxembourg \\
  }

\def\month{08}  
\def\year{2026} 
\def\openreview{\url{https://openreview.net/forum?id=6W3pFEQXZc}}
\AtBeginEnvironment{tabular}{\scriptsize}

\begin{document}
\maketitle

\begin{abstract}
Composed Image Retrieval (CIR) is an emerging paradigm in content-based image retrieval 
that enables users to formulate compositional queries 
by combining a reference image with an auxiliary modality, usually text-based. 
This approach supports fine-grained search where the target image shares structural elements with the user-provided image 
while incorporating the modifications specified by the auxiliary text. 
Conventional CIR methods rely on multimodal fusion to combine visual and textual features into a joint query embedding, 
which requires training modules that align composed queries with the targets. 
In this work, we propose \pefuse (for pseudo-fusion), a training-free framework 
that leverages pretrained Diffusion Models and Multimodal Large Language Models 
to bridge modalities via generative conversion. 
We introduce two novel strategies: uni-directional and bi-directional conversion, 
which convert CIR into four single-modality retrieval problems. 
These methods reformulate CIR as either intra-modal or cross-modal single-query retrieval tasks, 
bypassing the need for dedicated task-specific training. 
Extensive experiments on standard benchmarks demonstrate that converting CIR into text-to-image retrieval tasks 
is more effective than alternative conversion strategies, 
achieving competitive or superior performance compared with state-of-the-art methods, 
while maintaining high flexibility thanks to replaceable components of the conversion pipeline. 
These results highlight the effectiveness of the pseudo-fusion paradigm for zero-shot CIR.
Our code is publicly available at: \url{https://github.com/StevenXuf/PeFuse4CIR}.
\end{abstract}
\section{Introduction}
\label{sec:intrudction}

Traditional content-based image retrieval systems allow users to submit image-based queries,
bridging the so-called \emph{semantic gap}~\citep{Smeulders00}.
This constraint hinders their ability to accommodate nuanced search intents that are inherently multimodal. 
Composed Image Retrieval (CIR) addresses this limitation by enabling users to formulate a query 
using a reference image coupled with an auxiliary modality to specify desired modifications.
This approach facilitates fine-grained retrieval,
which is particularly valuable in domains such as e-commerce \citep{combiner}, digital asset management \citep{eufcc}, and creative design \citep{cir_survey}.

CIR introduces a number of challenges. 
An effective retrieval system must not only comprehend the individual modalities but also model their compositional semantics, 
capturing how the modification alters the meaning of the reference image. 
A prevalent solution involves multimodal feature fusion, 
in which visual and textual representations are integrated into a unified embedding prior to retrieval. 
Although recent advances in deep Transformer-based architectures~\citep{transformer, vit} 
have improved cross-modal alignment and compositional reasoning, 
the majority of existing methods rely heavily on dedicated training on large-scale, annotated CIR datasets. 
To alleviate these restrictions, researchers either synthesize triplet datasets~\citep{coalign, cig, context_cir} 
or rely on existing image-text pairs~\citep{hycir} to train models. 
This dependency limits their scalability and adaptability to significant domain shifts, 
for which zero-shot CIR (ZS-CIR) is regarded as an effective solution.

In this work, we investigate \textit{training-free pseudo-fusion} strategies for ZS-CIR 
that circumvent the need for additional task-specific fusion training. 
We propose to pseudo-fuse the multimodal query through \textit{uni-directional} and \textit{bi-directional conversion} techniques, 
leveraging recent advancements in Diffusion Models~\citep{diffmodels, ddim, latentdiff} 
and Multimodal Large Language Models (MLLMs)~\citep{llava, llama3, qwen2_5}. 
The uni-directional approach reformulates the multimodal-query CIR task into a standard uni-modal query problem
by converting the reference image and text modification into a synthesized image or a composed description. 
The bi-directional approach extends this by additionally converting the candidate images into text, 
enabling a text-based matching paradigm. 
These strategies facilitate flexible adaptation of existing, off-the-shelf retrieval systems 
without requiring architectural modifications, fine-tuning, or any training.

Extensive experiments on standard CIR benchmarks demonstrate that converting CIR tasks to text-to-image tasks 
via the proposed framework achieves competitive or superior performance compared to the state-of-the-art (SOTA) models. 
Our findings underscore the significant potential of training-free approaches in compositional retrieval. 
To the best of our knowledge, this is the first work to systematically explore and benchmark modality conversion strategies 
utilizing both Diffusion models and MLLMs for ZS-CIR.
In summary, our contributions are as follows:
\begin{itemize}
    \item Versatile training-free pseudo-fusion strategies for ZS-CIR 
    that seamlessly convert multimodal queries into a single modality, 
    enabling compatibility with existing retrieval models without task-specific fine-tuning.
    \item A systematic study and comprehensive benchmarking 
    of both uni-directional and bi-directional modality conversion paradigms for CIR using Diffusion models and MLLMs.
    \item A quantitative analysis to elucidate the relationship between CIR performance 
    and key hyperparameters of both MLLMs and Diffusion models, 
    as well as  a component-based latency analysis when utilizing MLLMs and Diffusion models to reframe CIR tasks to single-modality retrieval tasks.
    \item Empirical evidence that reformulating CIR as text-to-image retrieval is (for now) more effective than other tasks, 
    underscoring the substantial opportunity to further enhance image generation quality through Diffusion models.
\end{itemize}
\section{Related Work}
\label{sec:related_work}

Early CIR approaches like TIRG~\citep{tigr} relied on joint embedding spaces 
trained with contrastive objectives~\citep{infonce, simclr, moco}, 
where the fused image–text representation was directly compared against candidate image embeddings.
Subsequent Transformer-based methods~\citep{align, blip, blip2}, 
pretrained on large-scale Vision–Language datasets, achieved stronger cross-modal alignment.
Building on this foundation, Combiner~\citep{combiner} leverages CLIP~\citep{clip} 
to compute integrated features from reference images and accompanying textual descriptions.

A notable line of work builds upon the idea of representing images as pseudo-word tokens within a text sequence. 
Inspired by Textual Inversion~\citep{textualinversion}, 
methods such as SEARLE~\citep{searle}, Pic2Word~\citep{pic2word}, and LinCIR~\citep{lincir} 
map reference images into token embeddings that can be processed by language models, 
achieving SOTA performance through joint training. 
Other approaches like CLIP4CIR~\citep{clip4cir} 
introduce learnable fusion operators to better capture compositional semantics.
More recently, the generative capabilities of Diffusion models~\citep{diffmodels, ddim, latentdiff} have also been explored for CIR. 
For example, CIG~\citep{cig} uses a pretrained textual inversion network 
to convert a reference image into tokens that can be employed for target image generation and fused with the query. 

Despite these advances, the use of LLMs/MLLMs remains more prevalent. 
For example, DQU-CIR~\citep{dqu_cir} fuses unified textual and visual information extracted 
via LLMs or VLMs. 
Notably, HyCIR~\citep{hycir} enhances model training by incorporating contrastive learning on synthetic triplets, 
demonstrating the efficacy of synthetic supervision. 
Further advancing this approach, \cite{improvedcir} scales both negative and positive samples for contrastive learning using an MLLM.
Furthermore, MRA-CIR~\citep{mra_cir} circumvents error-prone intermediate text generation 
by using a Multimodal Reasoning Agent to directly construct high-quality triplets from unlabeled images. 
Similarly, CoLLM~\citep{collm} mitigates data scarcity by synthesizing training triplets from image-caption pairs 
using LLMs, enabling multimodal fusion.

A common characteristic of the aforementioned methods is their reliance on ad-hoc training, 
either on synthetic triplets or existing image-text datasets. 
Although learnable fusion methods achieve strong performance, 
their requirement for task-specific training limits flexibility and generalizability to new domains or modalities. 
To generalize to new datasets, training-free CIR methods have been proposed.
CIReVL~\citep{cirevl} uses an LLM to refine VLM-generated captions
by incorporating the text modification to perform text-to-image retrieval. 
Similarly, WeiMoCIR~\citep{weimocir} employs an MLLM to caption candidate target images. 
Moreover, LDRE~\citep{ldre} leverages LLMs to generate multiple captions that capture diverse semantic aspects of reference images conditioned on modifications, and subsequently assigns weights to these generated captions to enhance image retrieval performance.
Another relevant framework, ImageScope~\citep{imagescope}, unifies various language-guided image retrieval tasks 
into a text-to-image retrieval task using descriptions generated by an MLLM. 
However, it relies on multiple models in stages, 
leading to cumulative compute and increased inference time. 
In contrast, we use only either a single MLLM or Diffusion model, rather than multiple LLMs or MLLMs, resulting in a simpler yet effective pipeline for reformulating CIR to other tasks.

Unlike these prior works, which often involve multi-stage text generation or ensembles of multiple models, 
we reformulate CIR into four high-level tasks and systematically benchmark them using only one model where possible, 
including conversion to one-query intra-modal or cross-modal retrieval tasks, as either text-based or image-based retrieval. 
By leveraging pretrained Diffusion models and MLLMs, 
our approach offers a flexible, modular, and plug-and-play solution for ZS-CIR.
\section{Methodology}
\label{sec:methods}

A CIR query specifies its intent jointly through a reference image and a textual modification,
while the target images can be searched in a provided image gallery based on such queries. 
Instead of fusing the respective embeddings of each modality from the composed query as in prior work, our objective is to transform multimodal queries into a unified uni-modal representation, either as a Diffusion-generated image or an MLLM-generated textual description. 
This reformulation enables existing retrieval models to match the query against either target images or their corresponding MLLM-generated textual descriptions.

\subsection{Task Formulation}

Formally, let $\mathcal{I}$ and $\mathcal{T}$ be the image and text spaces, respectively. 
Let
\(
\mathcal{G}=\{I_{tar}^j\}_{j=1}^{N}\subset\mathcal{I}
\)
denote an image gallery. 
For CIR tasks, given a paired query \(q_i=(I_{ref}^{i},T_i), i=1, \dots, Q\), consisting of a reference image $I_{ref}^i\in\mathcal{I}$ and a text modification $T_i\in\mathcal{T}$, the goal is to retrieve target images from the image gallery $\mathcal{G}$ that are visually similar to $I_{ref}^i$ while satisfying the semantic requirements specified by $T_i$, based on a similarity metric:
\begin{equation}
    f(I_{ref}^i,T_i,\mathcal{G})
    = \operatornamewithlimits{argsort}_{I_{tar}^j\in\mathcal{G}}
      s(q_i,I_{tar}^j),
    \label{eq:cir_task}
\end{equation}
where \(s_{ij}:=s(q_i,I_{tar}^j)\) is the similarity score between the query and gallery images
and \texttt{argsort} is the sorting operator that ranks the scores by descending order. 
To use pretrained retrieval models, we convert the multimodal query $q_i$ into uni-modal interpretable representations, 
either text via MLLMs or images via Diffusion models.
To rank the results, we use cosine similarity,
which has been shown to provide a robust measure 
for multimodal representation alignment in cross-modal retrieval~\citep{Xu24align}.
The most relevant candidate images in the image gallery $I_{tar}^j\in\mathcal{G}$ 
for a given composed query $q_i$ are retrieved based on:
\begin{equation}
    s_{ij}
    =
    \cos\left(\Psi(q_i),\Psi(I_{tar}^j)\right)=\frac{\Psi(q_i)^{\top}\Psi(I_{tar}^j)}{\|\Psi(q_i)\|_2\,\|\Psi(I_{tar}^j)\|_2},
\end{equation}
where $\Psi(\cdot)$ is a pretrained retrieval model that takes both image $I\in\mathcal{I}$ and text $T\in\mathcal{T}$ as inputs, 
and outputs the extracted embeddings $\Psi(I)\in\mathbb{R}^m$ and $\Psi(T)\in\mathbb{R}^m$ in a shared embedding space $\Omega$. 

\subsection{Pseudo Fusion Strategies for Composed Queries}

\begin{figure}[!ht]
    \centering
    \includegraphics[width=\linewidth, keepaspectratio=true]{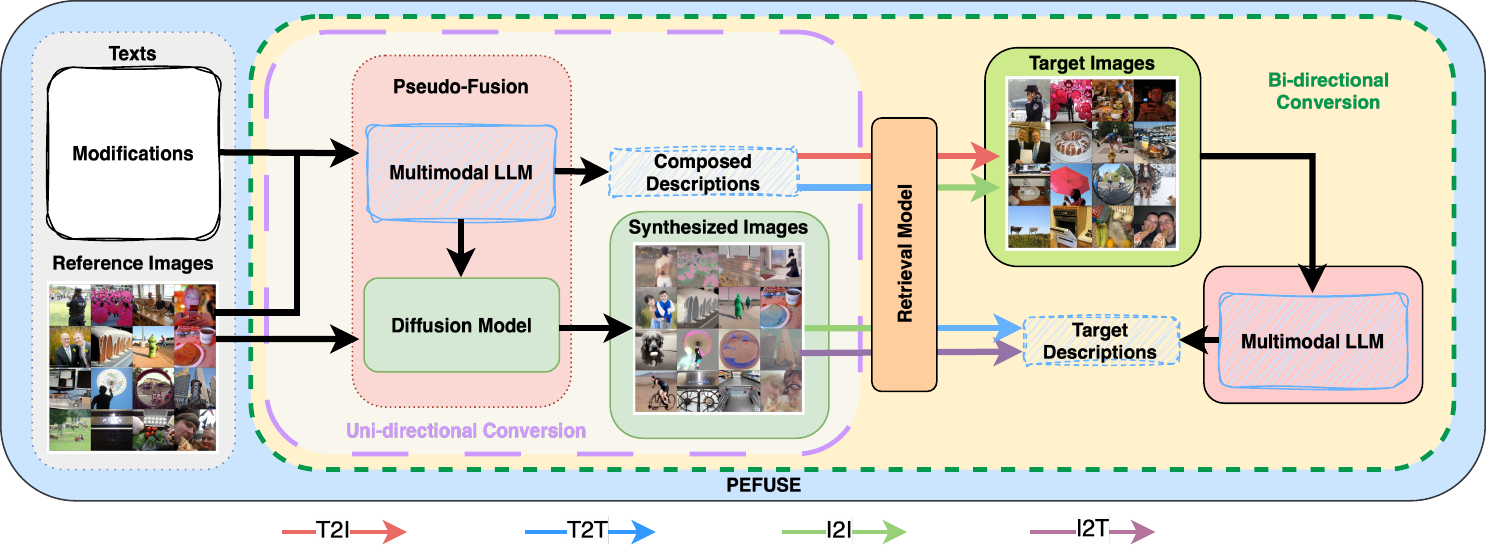}
    \caption{Training-free pseudo-fusion framework for Composed Image Retrieval.
    Dashed green box indicates bi-directional conversion, while the dashed purple box is uni-directional conversion.
    T2I: text-to-image; T2T: text-to-text; I2T: image-to-text; I2I: image-to-image. 
    Black arrows indicate the modality conversion flow.}
    \label{fig:pipeline}
\end{figure}

As depicted in \autoref{fig:pipeline}, our method employs a dual-strategy pipeline. 
The \emph{uni-directional conversion} (\textcolor{Thistle}{magenta} dashed box) 
facilitates retrieval by projecting the query into a target modality, 
either by using an MLLM to generate descriptive text from images and modifications 
or by using a Diffusion model to generate images from reference images and MLLM-generated text.
Specifically, we use MLLMs to convert reference images plus corresponding modifications into text 
or Diffusion models to generate alternative target images from such composed queries.
The \emph{bi-directional conversion} (\textcolor{ForestGreen}{green} dashed box) extends this by subsequently using the MLLM 
to also project target images into the textual modality, enabling a text-based retrieval process.
Namely, we additionally generate text based on target images, and then match with generated text or images from uni-directional conversion.

\paragraph{Uni-directional Conversion} For uni-directional conversion, we use an MLLM $h(\cdot)$ to generate textual descriptions $T_h$ 
based on an arbitrary combination of images $I_{ref}^i$ and text $T_i$ given the proper dataset-specific prompt $p_d$, i.e., $q_i:=T_h^i=h(I_{ref}^i,T_i,p_d)\in\mathcal{T}$,
while we use a Diffusion model $g(\cdot)$ to generate image $I_g$ based on both images $I_{ref}^i$ and text $T_i$, i.e., $q_i:=I_g^i=g(I_{ref}^i,T_h^i)\in\mathcal{I}$.
We feed the generated text $T_h^i$ to retrieval model $\Psi$
to compute cosine similarity with respect to all candidate images:
$s_{ij}=\cos\left(\Psi(T_h^i),\Psi(I_{tar}^j)\right)$, 
and similarly for synthesized images $I_g^i$: $s_{ij}=\cos\left(\Psi(I_g^i),\Psi(I_{tar}^j)\right)$, where $j\in\{1,\dots,N\}$. 
Through this method, we reformulate CIR into two types of image retrieval tasks: text-to-image and image-to-image as in the pink box of \autoref{fig:pipeline}.

\paragraph{Bi-directional Conversion} For bi-directional conversion, we additionally convert target images $I_{tar}^j$ into text via MLLMs given another dataset-specific prompt $q_d$: $T_h^j=h(I_{tar}^j,q_d)$ 
and compute the cosine similarity based on previously generated modalities,
either by $s_{ij}=\cos\left(\Psi(T_h^i),\Psi(T_h^j)\right)$ or $s_{ij}=\cos\left(\Psi(I_g^i),\Psi(T_h^j)\right)$ for $j\in\{1,\dots,N\}$, 
which reformulates the CIR tasks into text-to-text retrieval and image-to-text retrieval tasks in the green box of \autoref{fig:pipeline}, respectively.

We summarize \pefuse in \autoref{alg:pefuse}.
Based on the computed scores, we rank the candidates in descending order and return the top-$k$ candidate image IDs for performance evaluation. 
We compute the commonly used retrieval metrics according to the ground-truth labels and the returned top-$k$ candidates for each query on the standard official splits of datasets, and report the results.
Our framework is termed pseudo-fusion as it relies on generative models to synthesize new data (images or text) 
from pairs of elements within a triplet, thereby achieving an implicit fusion of modalities. 
This approach is distinct from typical early fusion paradigms, 
which explicitly combine modalities into intermediate embeddings. 
In contrast, our method directly generates coherent and interpretable data in a target modality, 
preserving latent semantics throughout the process.

\begin{algorithm}[!ht]
\caption{Training-Free Pseudo-Fusion (\pefuse) for Composed Image Retrieval}
\label{alg:pefuse}
\begin{algorithmic}[1]
\Requirements Retrieval model $\Psi(\cdot)$, MLLM $h(\cdot)$, Diffusion model $g(\cdot)$, prompts $p_d$ and $q_d$
\Require Reference image $I_{ref}^i$, modification text $T_i$, candidate image set $\mathcal{G}=\{I^j_{tar}\}_{j=1}^{N}$, retrieval mode $m$
\Ensure Ranked list of candidate images

\State Generate composed text description:
\[
T_h^i \leftarrow h(I_{ref}^i, T_i, p_d)
\]

\If{$m \in\{\textcolor{red}{\mathrm{T{\rightarrow}I}}, \textcolor{NavyBlue}{\mathrm{T{\rightarrow}T}}\}$}

    \If{$m = \textcolor{red}{\mathrm{T{\rightarrow}I}}$}
        \For{each candidate image $I^j_{tar} \in \mathcal{G}$}
            \State $s_{ij} \leftarrow \cos\left(\Psi(T_h^i), \Psi(I^j_{tar})\right)$
        \EndFor
        
    \ElsIf{$m = \textcolor{NavyBlue}{\mathrm{T{\rightarrow}T}}$}
        \For{each candidate image $I^j_{tar} \in \mathcal{G}$}
            \State Generate target description:
            \[
            T^j_h \leftarrow h(I^j_{tar}, q_d)
            \]
            \State $s_{ij} \leftarrow \cos\left(\Psi(T_h^i), \Psi(T^j_h)\right)$
        \EndFor
    \EndIf
        
\Else

    \State Generate synthesized target image:
    \[
    I_g^i \leftarrow g(I_{ref}^i, T_h^i)
    \]
    \If{$m = \textcolor{LimeGreen}{\mathrm{I{\rightarrow}I}}$}
        \For{each candidate image $I^j_{tar} \in \mathcal{G}$}
            \State $s_{ij} \leftarrow \cos\left(\Psi(I_g^i), \Psi(I^j_{tar})\right)$
        \EndFor

    \ElsIf{$m = \textcolor{Thistle}{\mathrm{I{\rightarrow}T}}$}
        \For{each candidate image $I^j_{\mathrm{tar}} \in \mathcal{G}$}
            \State Generate target description:
            \[
            T^j_h \leftarrow h(I^j_{tar}, q_d)
            \]
            \State $s_{ij} \leftarrow \cos\left(\Psi(I_g^i), \Psi(T^j_h)\right)$
        \EndFor
    \EndIf
\EndIf

\State Sort candidates in descending order of $s_{ij}$
\State \Return ranked candidate list
\end{algorithmic}
\end{algorithm}
\section{Experiments}
\label{sec:experiments}

We first describe the experimental setup, including the datasets and models used, 
and then present the main results for different conversion methods.
Note that multi-round conversion or using multiple models can also be incorporated into our pipeline for improved performance, 
whereas we minimize the number of models used to investigate which pathway 
is the most effective when converting the CIR task into a single-modality retrieval task.
Specifically, our experiments use either an MLLM or a Diffusion model for uni-directional conversion,
while both an MLLM and a Diffusion model are used for bi-directional conversion.

\subsection{Data and Models}

We employ four standard CIR benchmark datasets: Fashion-IQ~\citep{fashionIQ}, CIRR~\citep{cirr}, CIRCO~\citep{searle}, and GeneCIS~\citep{genecis}. 
Fashion-IQ is designed for interactive fashion image retrieval using natural language feedback, 
incorporating human-written relative captions and derived visual attributes. 
CIRR extends the scope to open-domain images with human-annotated modifying text, 
though it is known to contain a significant number of false negatives~\citep{searle}. 
To mitigate this issue, CIRCO provides multiple ground-truth images per query, 
with all images sourced from the MS-COCO~\citep{coco} dataset. 
In addition, GeneCIS measures models’ ability to adapt to a range of similarity conditions in terms of attributes and objects.
In line with standard evaluation protocols, we use the official splits of each dataset, along with the corresponding official target image candidate pools.
We report recall@$K$ for Fashion-IQ, CIRR, and GeneCIS, and mean average precision (mAP@$K$) for CIRCO, 
reflecting their respective annotation structures.
Note that predictions on the test splits of CIRR and CIRCO must be submitted to their respective evaluation servers for automatic assessment, as ground-truth labels are not publicly available.\footnote{CIRR server: \url{https://cirr.cecs.anu.edu.au}, CIRCO server: \url{https://circo.micc.unifi.it}} 
In contrast, Fashion-IQ and GeneCIS provide ground-truth labels, enabling the direct computation of the corresponding evaluation metrics.

For image synthesis based on textual and visual inputs, 
we utilize the pretrained \texttt{sdxl-instructpix2pix} model from the \texttt{diffusers} library, 
which is an instruction-tuned variant of InstructPix2Pix~\citep{instructpix2pix}, 
for its strong generative performance. 
Text generation is handled by \texttt{Qwen2.5-VL-7B-Instruct}, 
an advanced instruction-tuned SOTA MLLM based on Qwen~\citep{qwen2_5}. 
We later evaluate additional MLLMs and Diffusion models to compare retrieval performance 
and analyze computational overhead incurred by our pipeline in \autoref{sec:overhead}.

To evaluate the effectiveness of our approach, 
we employ several retrieval models as feature extractors, 
starting with models using a ViT-B/32 backbone: 
CLIP (ViT-B/32)~\citep{clip}, OpenCLIP (ViT-B/32, pretrained on \texttt{laion2b\_s34b\_b79k})~\citep{openclip}, and SigLIP2 (Base-Patch16)~\citep{siglip2}. 
To ensure a fair comparison with methods using the same backbone and to maintain maximum consistency in our pipeline, we use SigLIP2 with ViT-B-16 instead of ViT-B-32, since the available SigLIP2 ViT-B-32 checkpoint is designed for images at a resolution of $256\times256\,px$. We include SigLIP2 with ViT-B-16 as a complementary retrieval model, since ViT-B-16 has a model size comparable to the ViT-B-32 backbones used by CLIP and OpenCLIP, allowing us to control for backbone capacity as much as possible. 
Both CLIP and OpenCLIP adopt the softmax function in their loss functions, with OpenCLIP additionally benefiting from training on substantially larger datasets. In contrast, SigLIP2 improves semantic understanding over SigLIP~\citep{siglip}, which was trained with sigmoid-based contrastive loss.
We investigate scaling behavior of larger model variants in \autoref{sec:scaling}.

All retrieval models operate on input images resized to $224\times224\,px$, 
normalized to the $[0,1]$ range using model-specific normalization parameters. 
The Diffusion model requires $768\times768\,px$ inputs and produces outputs at the same resolution. 
For consistency, all images across datasets are resized to $768\times768\,px$ 
and normalized to $[0,1]$ prior to Diffusion processing.
During image generation with the Diffusion model, we use a guidance scale of $7.5$, 
an image guidance scale of $3.0$, and $30$ denoising steps for performance. 
For text generation with the MLLM, we set the temperature to $0.1$, top-$p$ to $0.9$, top-$k$ to $50$, 
and a maximum of $128$ new tokens to ensure deterministic outputs with reasonable length. 
A sensitivity study in \autoref{sec:sensitivity} examines the impact of varying these hyperparameters. 
Our implementation uses \texttt{PyTorch} on a single NVIDIA A100 with 40\,GB of memory.

\subsection{Uni-directional Conversion}
\label{sec:uni_conversion}

As introduced in \autoref{sec:methods}, uni-directional conversion can be implemented 
using either MLLMs or Diffusion models as pseudo-fusion methods for queries, 
converting CIR into single-modality image-based retrieval tasks.

The top rows (\pefuse (T$\rightarrow$I) and \pefuse (I$\rightarrow$I)) in \autoref{tab:fashioniq_perf} 
present the text-to-image and image-to-image retrieval results on the Fashion-IQ dataset. 
We mainly compare with zero-shot, training-free methods: CIReVL, LDRE, WeimoCIR, and ImageScope, 
while training-based methods are reported in \autoref{tab:fashioniq_perf_training} and \autoref{app:traning_methods}.
When reformulating CIR as a text-to-image retrieval task, both CLIP and OpenCLIP, which utilize the ViT-B/32 backbone, outperform CIReVL. 
Notably, OpenCLIP as the retriever surpasses the performance of most zero-shot methods 
that require model training, with results comparable to, though slightly lower than LinCIR, which uses a larger ViT-L/14 backbone, and to that of WeiMoCIR.

\begin{table}[!ht]
    \centering
    \caption{Performance (\%) on \textbf{Fashion-IQ} validation split using different retrieval models via \pefuse. 
    \textbf{Best} and \underline{second} best highlighted. $^{*}$: reproduced results; $^{\dag}$: results from original papers.
    Training-based baselines are reported in \autoref{tab:fashioniq_perf_training}  in \autoref{app:traning_methods}.
    } 
    \begin{tabular}[width=\linewidth]{llrr rr rr rr}
        \toprule
         \multirow{2}*{\bf Method}&\multirow{2}*{\bf Retrieval Model}&\multicolumn{2}{c}{\bf Shirt}&\multicolumn{2}{c}{\bf Dress}&\multicolumn{2}{c}{\bf Toptee}&\multicolumn{2}{c}{\bf Average}  \\
         \cmidrule(lr){3-4}\cmidrule(lr){5-6}\cmidrule(lr){7-8}\cmidrule(lr){9-10}
         &&R@10&R@50&R@10&R@50&R@10&R@50&R@10&R@50 \\
         
        \midrule
            CIReVL$^{*}$&CLIP (ViT-B/32)&18.40&30.82&14.25&30.45&18.00&34.33&16.88&31.87 \\
            LDRE$^{\dag}$&CLIP (ViT-B/32)&27.38&\underline{46.27}&19.97&41.84&27.07&48.78&24.81&45.63\\
            WeiMoCIR$^{\dag}$&OpenCLIP (ViT-B/32)&\textbf{29.20}&\textbf{47.15}&\textbf{26.23}&\textbf{46.31}&\textbf{34.17}&\textbf{55.99}&\textbf{29.86}&\textbf{49.82}\\
            ImageScope$^{\dag}$&CLIP (ViT-B/32)&24.29&37.49&18.00&35.20&24.99&41.41&22.42&38.03 \\
        \midrule
          \multirow{3}*{\pefuse (T$\rightarrow$I)}&CLIP &20.62&37.11&13.99&32.54&19.93&39.58&18.18&36.41 \\
          &OpenCLIP &\underline{28.30}&46.19&\underline{24.05}&\underline{44.11}&\underline{32.46}&\underline{53.94}&\underline{28.27}&\underline{48.08}\\
          &SigLIP2 &6.49&13.61&7.32&17.05&7.23&16.93&7.01&15.86\\
        \midrule
        \multirow{3}*{\pefuse (I$\rightarrow$I)}&CLIP&8.97&17.01&4.95&13.82&7.71&16.76&7.21&15.87 \\
          &OpenCLIP&14.28&24.48&10.33&22.43&12.69&24.37&12.43&23.76 \\
          &SigLIP2&15.00&26.39&9.25&20.93&13.77&25.01&12.67&24.11 \\
        \midrule
        \multirow{3}*{\pefuse (T$\rightarrow$T)}&CLIP &14.33&25.57&8.82&20.55&16.01&29.57&13.06&25.23 \\
          &OpenCLIP &16.39&26.96&13.23&27.92&19.39&34.33&16.34&29.74 \\
          &SigLIP2 &2.53&5.88&1.72&5.38&3.05&7.61&2.43&6.29 \\
        \midrule
        \multirow{3}*{\pefuse (I$\rightarrow$T)}&CLIP &10.67&20.93&5.16&16.41&8.94&20.03&8.26&19.12 \\
          &OpenCLIP &14.28&26.55&8.18&19.96&12.32&25.44&11.59&23.98 \\
          &SigLIP2 &8.61&17.27&5.59&15.33&8.94&19.34&7.72&17.31 \\
        \bottomrule
    \end{tabular}
    \label{tab:fashioniq_perf}
\end{table}

In general, when comparing \pefuse (T$\rightarrow$I) and \pefuse (I$\rightarrow$I), 
text-to-image retrieval demonstrates superior performance compared to image-to-image retrieval; 
the sole exception is observed with the SigLIP2 model. 
Across all retrieval models evaluated, we note a significant inconsistency in performance: 
SigLIP2 yields the weakest results for text-to-image retrieval, 
yet achieves the strongest performance for image-to-image retrieval. 
This disparity underscores the substantial differences in semantic understanding capabilities across retrieval models on varying tasks, 
highlighting their critical and impactful role in the effectiveness of CIR systems.

The image retrieval performance on the CIRR and CIRCO datasets 
is further detailed in \autoref{tab:cirr_perf} and \autoref{tab:circo_perf}, respectively.
Training-based baseline methods are reported in \autoref{tab:cirr_perf_training} and \autoref{tab:circo_perf_training}, with additional results provided in \autoref{app:traning_methods}.
On the CIRR dataset, for the text-to-image retrieval task, the CLIP model slightly underperforms compared to other methods while being notably better on CIRR subsets. 
In contrast, SigLIP2 and OpenCLIP achieve significantly stronger performance on both CIRR and its subsets, although they fall behind ImageScope in some cases. 
For the image-to-image task on CIRR, all retrieval models fall behind the established baselines, 
underscoring the superiority of reformulating CIR as a text-to-image rather than an image-to-image retrieval task.

\begin{table}[t]
    \centering
    \caption{Performance (\%) on \textbf{CIRR} test split using different retrieval models via \pefuse.
    \textbf{Best} and \underline{second} best highlighted. $^{*}$: reproduced results; $^{\dag}$: results from original papers; $-$: results not available.}
    Training-based baselines are reported in \autoref{tab:cirr_perf_training}  in \autoref{app:traning_methods}.
    \begin{tabular}{ll rrrrr rrr}
        \toprule
         \multirow{2}*{\bf Method}&\multirow{2}*{\bf Retrieval Model}&\multicolumn{5}{c}{\bf Recall}&\multicolumn{3}{c}{\bf Recall$_{\text{subset}}$}\\
         \cmidrule(lr){3-7}\cmidrule(lr){8-10}
         &&@1&@2&@5&@10&@50&@1&@2&@3 \\
        \midrule
         CIReVL$^{*}$&CLIP (ViT-B/32)&21.40&31.86&47.74&60.72&84.99&56.27&77.08&88.63 \\
         LDRE$^{\dag}$&CLIP (ViT-B/32)&25.69&\textemdash&55.13&69.04&89.90&60.53&80.65&90.70\\
         WeiMoCIR$^{\dag}$&OpenCLIP (ViT-B/32)&26.31&\textemdash&57.69&70.36&91.01&53.35&75.57&87.76\\
         ImageScope$^{\dag}$&CLIP (ViT-B/32)&\textbf{34.36}&\textemdash&60.58&71.40&88.41&\textbf{74.63}&\underline{87.93}&\underline{93.83} \\
        \midrule
         \multirow{3}*{\pefuse (T$\rightarrow$I)}
         &CLIP &22.58&32.96&49.81&63.59&87.47&63.28&82.27&91.76 \\
         &OpenCLIP &\underline{34.00}&\textbf{47.49}&\textbf{65.57}&\textbf{77.47}&\textbf{93.28}&\underline{71.59}&\textbf{88.39}&\textbf{95.04} \\
         &SigLIP2 &30.65&\underline{43.13}&\underline{60.60}&\underline{72.43}&\underline{91.35}&70.00&86.48&93.64 \\
        \midrule
         \multirow{3}*{\pefuse (I$\rightarrow$I)}& CLIP &2.77&9.78&23.06&35.42&64.58&29.90&52.19&70.96 \\
         & OpenCLIP &3.28&11.64&27.76&41.45&71.37&31.01&53.83&72.12 \\
         & SigLIP2 &3.78&12.72&28.39&42.12&70.07&32.43&54.15&71.35 \\
        \midrule
         \multirow{3}*{\pefuse (T$\rightarrow$T)}& CLIP&21.81&31.13&45.06&56.72&78.51&59.47&79.66&90.48 \\
         & OpenCLIP&30.65&42.96&58.99&70.68&89.25&69.45&86.00&93.81 \\
         & SigLIP2&16.07&24.19&37.40&47.81&68.82&52.77&72.39&84.65 \\
        \midrule
        \multirow{3}*{\pefuse (I$\rightarrow$T)}&CLIP&6.80&13.76&28.27&42.12&72.27&36.12&57.83&74.80 \\
        &OpenCLIP&7.40&15.86&31.71&45.16&74.29&36.10&58.72&75.47 \\
        &SigLIP2&7.52&14.12&26.82&39.13&67.49&34.41&56.63&74.41 \\
        \bottomrule
    \end{tabular}
    \label{tab:cirr_perf}
\end{table}

\begin{table}[t]
    \centering    
    \caption{Performance (\%) on \textbf{CIRCO} test split using different retrieval models via \pefuse.
    \textbf{Best} and \underline{second} best highlighted. $^{*}$: reproduced results; $^{\dag}$: results from original papers.
    Training-based baselines are reported in \autoref{tab:circo_perf_training}  in \autoref{app:traning_methods}.
    }
    \begin{tabular}{llrrrr}
        \toprule
         \bf Method&\bf Retrieval Model&\bf mAP@5&\bf mAP@10&\bf mAP@25&\bf mAP@50 \\
        \midrule
        CIReVL$^{*}$&CLIP (ViT-B/32)&10.36&10.70&11.88&12.47 \\
        LDRE$^{\dag}$&CLIP (ViT-B/32)&17.96&18.32&20.21&21.11\\
        ImageScope$^{\dag}$&CLIP (ViT-B/32)&\textbf{22.36}&\textbf{22.19}&\textbf{23.03}&\textbf{23.83} \\
        \midrule
         \multirow{3}*{\pefuse (T$\rightarrow$I)}& CLIP &11.12&11.47&12.86&13.61 \\
         &OpenCLIP&16.89&17.56&19.14&20.16 \\ 
         & SigLIP2&\underline{18.53}&\underline{19.68}&\underline{21.58}&\underline{22.62 }\\ 
        \midrule
         \multirow{3}*{\pefuse (I$\rightarrow$I)}& CLIP&2.39&2.57&3.08&3.37 \\
         & OpenCLIP&3.03&3.49&4.10&4.44 \\
         & SigLIP2&4.06&4.63&5.47&5.94 \\
        \midrule
         \multirow{3}*{\pefuse (T$\rightarrow$T)}& CLIP&7.71&7.97&8.76&9.26 \\
         & OpenCLIP&11.52&11.86&13.07&13.71 \\
         & SigLIP2&6.68&6.56&7.14&7.48\\
        \midrule
        \multirow{3}*{\pefuse (I$\rightarrow$T)}&CLIP&3.89&4.35&4.88&5.29 \\
        &OpenCLIP&4.10&4.62&5.26&5.74 \\ 
        &SigLIP2&3.65&3.84&4.40&4.78 \\
        \bottomrule
    \end{tabular}
    \label{tab:circo_perf}
\end{table}

A similar phenomenon is observed on the CIRCO dataset, 
where text-to-image retrieval outperforms most baseline methods except ImageScope, 
while all the baselines outperform the image-to-image conversion mode.
Specifically, text-to-image retrieval using CLIP performs competitively, 
exceeding training-free methods such as CIReVL, 
though it remains behind LinCIR and HyCIR, both of which employ a larger ViT-L/14 backbone and demand training. 
Notably, within the same task, using OpenCLIP and SigLIP2 with a ViT-B/32 backbone significantly improves the performance, 
and SigLIP2 surpasses most baseline methods by a considerable margin while being behind the top performer ImageScope. 
This indicates that employing a more powerful retrieval model can substantially enhance system performance. 
Conversely, experiments on image-to-image retrieval for CIRCO demonstrate inferior results, 
highlighting a need for improvement in Diffusion-based conversion methods.
We show results on GeneCIS in \autoref{tab:genecis}
and those of training-based methods in \autoref{tab:genecis_training} in \autoref{app:traning_methods}. 
From \autoref{tab:genecis}, we find text-to-image retrieval still performs the best while other conversion modes are still competitive with the baselines, which further corroborates our analysis on \pefuse (T$\rightarrow$I) and \pefuse (I$\rightarrow$I).

\begin{table}[!ht]
    \centering
    \caption{Performance (\%) comparison on \textbf{GeneCIS} test split using different retrieval models via \pefuse. \textbf{Best} and \underline{second} best highlighted. $^{\dag}$: results from original papers.
    Training-based baselines are reported in \autoref{tab:genecis_training}  in \autoref{app:traning_methods}.
    }
    \resizebox{\linewidth}{!}{
    \begin{tabular}{ll rrr rrr rrr rrr r}
        \toprule
         \multirow{2}*{\textbf{Method}}&\multirow{2}*{\textbf{Retrieval Model}}&\multicolumn{3}{c}{\textbf{Focus Attribute}}&\multicolumn{3}{c}{\textbf{Change Attribute}}&\multicolumn{3}{c}{\textbf{Focus Object}}&\multicolumn{3}{c}{\textbf{Change Object}}&\textbf{Average}  \\
         \cmidrule(lr){3-5}\cmidrule(lr){6-8}\cmidrule(lr){9-11}\cmidrule(lr){12-14}
         &&R@1&R@2&R@3&R@1&R@2&R@3&R@1&R@2&R@3&R@1&R@2&R@3&R@1\\
        \midrule
        CIReVL$^\dag$&CLIP (ViT-B/32)&17.90&29.40&40.40&\textbf{14.80}&25.80&35.80&14.60&24.30&33.30&16.10&27.80&37.60&15.90\\
        \midrule
        \multirow{3}*{\pefuse (T$\rightarrow$I)}&CLIP&\underline{18.20}&\underline{31.15}&41.85&14.54&25.76&35.89&13.67&25.26&34.90&\underline{17.40}&29.03&38.67&15.95\\
        &OpenCLIP&\textbf{19.55}&\textbf{31.95}&\textbf{43.70}&\underline{14.73}&\underline{26.28}&\textbf{36.51}&\underline{17.96}&\underline{28.27}&\underline{38.06}&16.73&\underline{30.97}&\underline{41.33}&\underline{17.22}\\
        &SigLIP2&17.00&29.35&39.65&14.63&\textbf{26.70}&\underline{36.17}&\textbf{18.27}&\textbf{29.34}&\textbf{38.27}&\textbf{19.59}&\textbf{31.58}&\textbf{42.24}&\textbf{17.37}\\
        \midrule
        \multirow{3}*{\pefuse (I$\rightarrow$I)}&CLIP&15.90&27.15&37.50&11.03&20.36&27.60&9.18&18.88&27.35&8.16&17.60&27.65&11.07\\
        &OpenCLIP&17.60&30.65&41.70&11.22&21.07&29.36&10.41&19.90&28.42&9.08&18.83&28.06&12.08\\
        &SigLIP2&17.95&29.15&40.00&11.17&21.69&30.21&10.20&19.49&27.91&9.90&20.71&30.71&12.31\\
        \midrule
        \multirow{3}*{\pefuse (T$\rightarrow$T)}&CLIP&16.55&28.80&39.65&10.18&20.45&29.02&14.85&24.44&33.27&12.30&23.98&32.60&13.47\\
        &OpenCLIP&16.60&30.90&\underline{41.95}&12.26&22.44&32.81&16.53&27.35&35.87&16.17&27.91&37.96&15.39\\
        &SigLIP2&14.60&25.75&35.65&9.42&17.95&25.95&11.84&20.00&28.37&11.02&19.64&29.44&11.72\\
        \midrule
        \multirow{3}*{\pefuse (I$\rightarrow$T)}&CLIP&17.45&30.70&41.00&10.94&21.07&30.45&10.97&19.74&28.16&11.58&20.82&29.64&12.74\\
        &OpenCLIP&17.45&30.45&41.75&10.84&21.40&31.20&11.84&21.89&30.77&11.22&21.07&30.26&12.84\\
        &SigLIP2&17.40&29.60&39.80&9.90&20.69&31.11&11.22&20.61&28.78&10.51&19.80&29.54&12.26\\
        \bottomrule
    \end{tabular}
    }
    \label{tab:genecis}
\end{table}

The experimental results across all datasets demonstrate that the framework is effective for ZS-CIR tasks,
despite its simplicity and training-free nature.
Although CLIP has been widely adopted in previous studies, our results indicate that it is a suboptimal choice for CIR systems compared to OpenCLIP, 
emphasizing the importance of choosing appropriate  retrieval models for CIR performance.
We further note that methods which separately generate captions via an image captioner 
and then combine them with modification text using an LLM (e.g., CIReVL) 
can be effective for simple images, such as fashion items. 
However, in complex scenarios like those in CIRCO, which involve a large pool of candidate images (123K), 
such pipelines often fail to adequately capture the intricate interactions between reference images and textual modifications, 
leading to inferior retrieval performance. 
Consequently, employing an MLLM proves to be both sufficient and less error-prone, 
outperforming lengthy chained pipelines for complex image retrieval tasks.
Additionally, applying methods from LDRE for MLLM-generated captions can further improve image retrieval performance.

\subsection{Bi-directional Conversion}
\label{sec:bi_conversion}

In addition to leveraging an MLLM to fuse the information from reference images 
and their corresponding modification text into textual descriptions, 
the same model is employed to generate descriptive captions for target images. 
This methodology effectively reformulates the CIR task into a text-to-text retrieval problem. 
Together with generated images via Diffusion models, the CIR task is reframed as an image-to-text retrieval task.
Previously, we reformulated the CIR task as either image-to-image retrieval via a Diffusion model or text-to-image retrieval via an MLLM.
Now, we show the results when reformulating CIR to single-modality text retrieval tasks, i.e., \pefuse (T$\rightarrow$T) and \pefuse (I$\rightarrow$T), 
with additional conversion on target images using an MLLM.
We report results in the bottom rows of \autoref{tab:fashioniq_perf}, \autoref{tab:cirr_perf}, \autoref{tab:circo_perf}, and \autoref{tab:genecis} respectively.

We observe that reformulating CIR as a text-to-text retrieval task 
generally yields stronger performance compared to image-to-text retrieval. 
Furthermore, 
both text-to-text and image-to-text retrieval underperform relative to text-to-image retrieval. 
However, image-to-text retrieval generally surpasses image-to-image performance. 
Overall, using Diffusion-generated images as queries to retrieve either original target images or MLLM-generated textual descriptions, i.e., \pefuse (I$\rightarrow$I) and \pefuse (I$\rightarrow$T), consistently underperforms methods that use MLLM-generated textual descriptions as queries, namely \pefuse (T$\rightarrow$I) and \pefuse (T$\rightarrow$T). One possible explanation is that images synthesized by Diffusion models  often contain artifacts that distinguish them from natural images. 
See \autoref{app:qualitative} for generated examples.
These artifacts can be captured by the image encoder of the retrieval model, thereby increasing the modality gap and degrading retrieval performance, whereas text generated by MLLMs retains more semantically meaningful information.
Furthermore, as reported in \cite{crossthegap}, models trained with cross-modal contrastive objectives generally perform better on cross-modal retrieval tasks than on intra-modal retrieval tasks (e.g., image-to-image or text-to-text retrieval). 
This is because cross-modal contrastive learning explicitly aligns representations across modalities but does not impose strong constraints on relationships within the same modality. 
Moreover, incorporating additional generative models into the retrieval pipeline may introduce cumulative errors at multiple stages, which can further contribute to performance degradation.
These findings further demonstrate that reformulating CIR as a text-to-image retrieval task 
is generally more effective than other conversion strategies under the same settings when employing CLIP-style retrieval models.

\subsection{Scaling Laws}\label{sec:scaling}

Having evaluated the performance of different conversion strategies 
using retrieval models with a ViT-B/32 backbone in \autoref{sec:uni_conversion} and \autoref{sec:bi_conversion}, 
a subsequent question arises regarding the potential benefits of larger backbone architectures. 
To investigate this, we assess the performance of OpenCLIP, 
which is selected for its superior overall performance among the three retrieval models, 
using ViT backbones of varying sizes across all datasets for text-to-image retrieval tasks due to their superior performance compared to other conversion modes. 
The overall results are presented in \autoref{tab:scaling_law}, 
with detailed category-specific results for Fashion-IQ and CIRR subsets 
provided in \autoref{tab:scaling_law_fashioniq_cat} in \autoref{app:scaling_law_ext},
and category-specific results of GeneCIS can be found in \autoref{tab:scaling_law_genecis_cat} in \autoref{app:scaling_law_ext}.

\begin{table}[!ht]
    \centering
    \caption{Scaling laws when using different pretrained backbones for OpenCLIP for text-to-image retrieval on each benchmark.}
    \resizebox{\linewidth}{!}{
    \begin{tabular}{l rr rrrrr rrrr r}
    \toprule
         \multirow{2}*{\bf Backbone}&\multicolumn{2}{c}{\bf Fashion-IQ}&\multicolumn{5}{c}{\bf CIRR}&\multicolumn{4}{c}{\bf CIRCO}&\textbf{GeneCIS}  \\
    \cmidrule(lr){2-3}\cmidrule(lr){4-8}\cmidrule(lr){9-12}\cmidrule(lr){13-13}
    &R@10&R@50&R@1&R@2&R@5&R@10&R@50&mAP@5&mAP@10&mAP@25&mAP@50&R@1\\
    \midrule
         ViT-L/14 (\texttt{laion2b\_s32b\_b82k})&29.49&48.37&36.17&50.07&67.13&78.72&94.17&21.69&22.99&25.07&26.14&16.85 \\
         ViT-H/14 (\texttt{laion2b\_s32b\_b79k})&30.40&50.09&38.55&52.02&69.49&80.29&94.29&23.78&24.78&27.10&28.24&17.50 \\
         ViT-g/14 (\texttt{laion2b\_s34b\_b88k})&30.15&50.12&38.41&52.15&70.15&80.27&94.58&22.62&23.93&26.39&27.42&17.39 \\
         ViT-bigG/14 (\texttt{laion2b\_s39b\_b160k})&30.44&49.49&40.41&54.63&71.13&81.06&94.89&25.03&26.63&29.17&30.28&17.99 \\
    \bottomrule
    \end{tabular}
    }
    \label{tab:scaling_law}
\end{table}

As shown in \autoref{tab:scaling_law}, we observe a general trend of improving performance 
for the text-to-image retrieval task as the backbone size increases, 
although performance fluctuates across specific model sizes. 
Notably, on Fashion-IQ, Recall@10 decreases and Recall@50 saturates 
when using the ViT-g/14 backbone. 
Performance on CIRR also improves consistently with model scale, 
with the exception of a slight decrease in Recall@1 and Recall@10 for ViT-g/14. 
A similar performance drop with ViT-g/14 is observed on the CIRCO dataset.
A similar trend holds for GeneCIS.
Furthermore, this trend of scaling benefits extends beyond text-to-image retrieval; 
larger models consistently achieve superior performance on image-to-image, text-to-text, and image-to-text retrieval tasks 
as well when using our proposed methods to reformulate CIR tasks. 

\subsection{Sensitivity Analysis}
\label{sec:sensitivity}

We study how the model performance is affected by varying values for hyperparameters of the MLLM and the Diffusion model used. 
We quantify the relationship between CIR performance and these hyperparameters via experimental studies.

\subsubsection{Multimodal Large Language Models}

For MLLMs,
temperature controls the determinism of the LLM when generating tokens, 
Top-$p$ is the cumulative probability threshold used when selecting tokens up to probability $P$, 
and Top-$k$ sampling limits the model to consider only the k most likely tokens at each step. 
For consistency with previous experiments, we use $0.1$ for temperature, $0.9$ for top-$p$, and $50$ for top-$k$ as the base combination 
and only alter one parameter while the others are fixed.
For example, when we investigate how temperature affects the retrieval performance, 
we fix top-$p$ to 0.9 and top-$k$ to 50, and change values for temperature in the range (0, 1).
We use the average of mAP@$k$ for the y-axis, where $k\in\{1, 5, 10,25, 50\}$.

\begin{figure}[!ht]
    \centering
    \includegraphics[width=\linewidth]{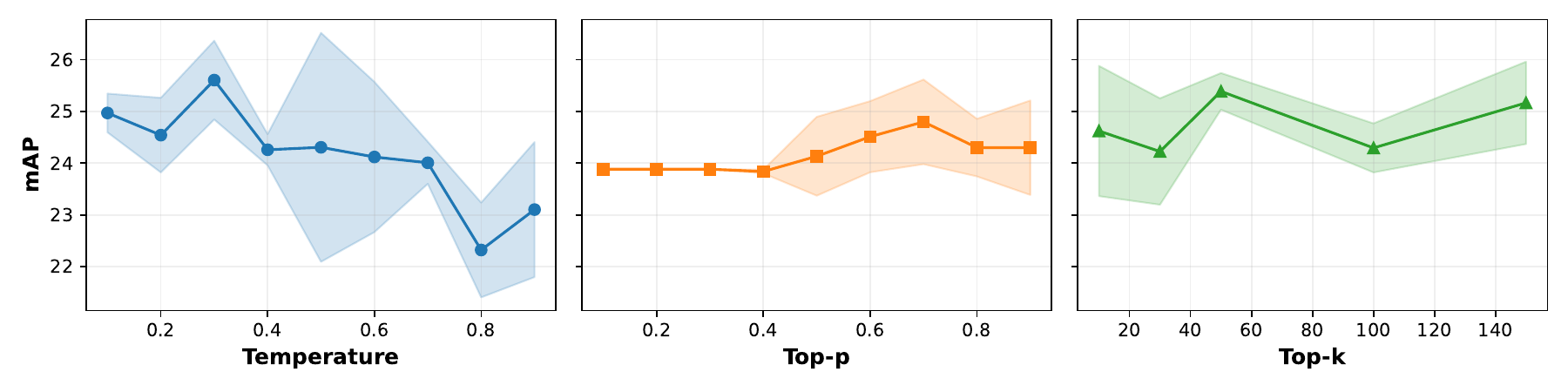}
    \caption{CIR performance (\%) of \pefuse (T$\rightarrow$I) via the MLLM under varying values for hyperparameters on the CIRCO validation split using OpenCLIP with a ViT-B/32 backbone for text-to-image retrieval over three runs. Shaded regions indicate the standard deviation.}
    \label{fig:ablation_llm_circo}
\end{figure}

We investigate the performance of different values of hyperparameters on the CIRCO validation dataset
(due to its manageable size and diverse nature of images) 
using OpenCLIP as the retrieval model in \autoref{fig:ablation_llm_circo}. 
From \autoref{fig:ablation_llm_circo}, we observe that retrieval performance 
is more strongly affected by temperature instead of top-$p$ or top-$k$. 
At higher temperatures, the model will produce more diverse tokens, 
which hurts retrieval performance when matching with images, 
whereas the performance of top-$p$ and top-$k$ is very stable overall. 
Overall, performance under the same set of hyperparameters shows slight differences, especially for top-$p$ and top-$k$.
The figure indicates that a lower temperature and a moderate top-$p$ with higher top-$k$ would result in better retrieval performance.

\subsubsection{Diffusion Models}

In the inference process of Diffusion models, 
a strong correlation exists between key hyperparameters and the properties of the synthesized output. 
Specifically, a larger number of inference steps correlates strongly with enhanced photorealism of the generated images. 
Furthermore, increasing the image guidance scale elevates the fidelity of the output to a given reference image. 
Conversely, a higher text guidance scale promotes stricter adherence to the input text prompt, 
often at the expense of output diversity. 
We use $7.5$ for guidance scale, $3.0$ for image guidance scale, 
and $30$ for inference steps as the base combination and only change one hyperparameter and fix the rest.

\begin{figure}[!ht]
    \centering
    \includegraphics[width=\linewidth]{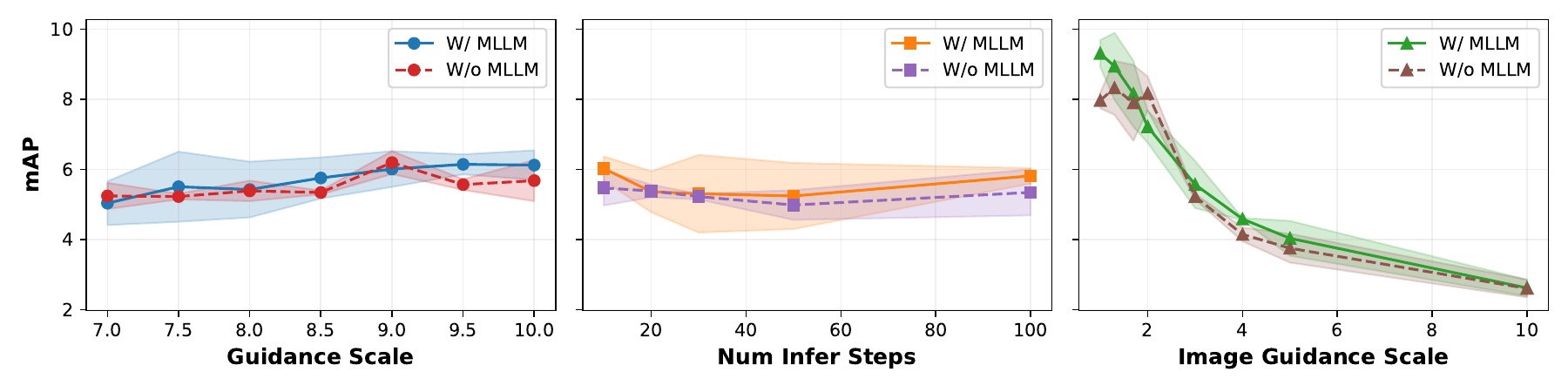}
    \caption{CIR performance (\%) of \pefuse (I$\rightarrow$I) via Diffusion models \emph{with} and \emph{without} using composed descriptions from the MLLM on CIRCO validation split using OpenCLIP with a ViT-B/32 backbone for image-to-image retrieval over three runs.
    Shaded regions indicate the standard deviation.}
    \label{fig:ablation_df_circo}
\end{figure}

We show the retrieval performance when using different values for hyperparameters in Diffusion models 
on the CIRCO validation dataset \emph{with} and \emph{without} MLLMs in \autoref{fig:ablation_df_circo}.
From \autoref{fig:ablation_df_circo}, we observe that using MLLM-generated descriptions 
to generate images improves the retrieval performance for all three hyperparameters, 
which validates the importance of using MLLM-generated descriptions as prompts instead of the original captions from datasets. 
We also show qualitative results with and without MLLMs for Diffusion models in \autoref{app:qualitative}.
From the synthesized images, we observe more distinguishable artifacts when directly using raw modifications from the CIRCO dataset.

The performance gap between using and not using MLLMs increases with stronger guidance scales and more inference steps
while the performance gap narrows with larger values for image guidance scale.
Across the three hyperparameters, image guidance scale has the most significant impact on retrieval performance, 
and higher values cause much worse performance. 
With a larger image guidance scale, the generated images would be more similar to the original input images than to the intended target images, 
thus deviating from target images and leading to worse performance.
This indicates the importance of low values for image guidance scale to achieve good performance 
when reformulating the CIR task as an image-to-image retrieval task. 

We also noticed that retrieval performance increases slightly as the number of inference steps increases, 
as more inference steps produce more photorealistic images but require more time. 
For time efficiency, a moderate number of inference steps should be sufficient, as indicated from the figure.
Generally speaking, the retrieval performance when employing Diffusion models via \pefuse (I$\rightarrow$I) is consistent across different runs.
Finally, we acknowledge that carefully tuning values for hyperparameters is labor-intensive, 
as different datasets and models might perform differently for the same setting of hyperparameters.
\section{Ablation and Latency Analysis}
\label{sec:overhead}

To confirm how the MLLMs and Diffusion models impact or contribute to downstream retrieval performance, we replace the \texttt{Qwen2.5-VL-7B-Instruct} and \texttt{SDXL-InstructPix2Pix} with other pretrained models and measure the metric of interest, which serves as an ablation of the components in our pipeline.
To quantify the latency incurred and obtain a clearer picture of the time efficiency of our pipeline, we also analyze the computational overhead of the proposed framework for the CIR task. 
We focus on \pefuse (T$\rightarrow$I) and \pefuse (I$\rightarrow$I), as they involve both MLLMs and Diffusion models. 
The former leverages MLLMs to compose target descriptions, while the latter employs Diffusion models, 
either guided by MLLM-generated descriptions or by raw modifications, to generate target images. 
It is worth noting that both settings involve image-retrieval-based uni-directional conversion, 
without converting the generated target images into text.

For consistency, we use the same OpenCLIP (ViT-B-32) in \autoref{sec:experiments} on CIRCO validation split for both tasks across experiments. 
Specifically, we additionally employ the \texttt{Qwen2.5-VL-3B-Instruct} to examine the latency incurred by MLLM scaling, 
and the similarly sized \texttt{LLaVA-1.5-7B-hf}~\citep{llava1_5} to investigate the impact of different training strategies. 
Notably, \texttt{Qwen2.5-VL-7B-Instruct} is trained with large-scale joint vision-language pretraining, 
whereas \texttt{LLaVA-1.5-7B-hf} aligns visual features with those of a frozen LLM.
In addition, we include \texttt{SDXL-Turbo}~\citep{sdxl_turbo} with a comparable parameter scale to \texttt{SDXL-InstructPix2Pix}, 
but supports faster inference, thereby illustrating the trade-off between inference speed and output quality.

We report the GPU memory consumption of each generative model considered during inference under \texttt{torch.bfloat16} precision, 
the average inference time per query, and the entire pipeline time, 
defined as the average end-to-end processing time per sample, including: 
dataset loading, model loading, data generation, feature extraction, and performance evaluation.
To isolate the effect of different components on retrieval performance, we also consider three retrieval settings as baseline evaluations: (1) reference image-to-target image, (2) modification text-to-target image, and (3) reference image plus modification text-to-target image retrieval. 
These settings are deterministic and performed via representation extraction using OpenCLIP. 
For all three settings, we directly apply retrieval models to extract respective representations for both image and text modalities, and compute the cosine similarity between them.
Particularly, representations of reference images and those of corresponding modifications are added and normalized for reference image plus modification text-to-target image retrieval.
Note that, for the baseline evaluations, the pipeline time directly reflects retrieval time, since no generation step is involved. 
For \pefuse (T$\rightarrow$I) with multiple MLLMs, caption-generation cost is captured by the inference time. 
For \pefuse (I$\rightarrow$I) with Diffusion models, the inference time represents the per-query image-generation time for both \texttt{SDXL-InstructPix2Pix} and \texttt{SDXL-Turbo}. 
When Diffusion models are chained with \texttt{Qwen2.5-VL-7B-Instruct}, the reported inference time includes both stages: first, generating the target-image description with the MLLM, and second, using that description as the refined prompt together with the reference image to generate an image with a Diffusion model.
We adopt the same metric (the average mAP across $\{1, 5, 10, 25, 50\}$), a batch size of 8 (for limited memory), 
and the same MLLM and Diffusion sampling hyperparameters as in \autoref{sec:sensitivity}.
For \texttt{SDXL-Turbo} we use 60 denoising steps and an image guidance scale of 0.5, resulting in $60 \times 0.5 = 30$ effective inference steps.\footnote{\url{https://huggingface.co/stabilityai/sdxl-turbo}}

We show the compute information in \autoref{tab:overhead}.
In \autoref{tab:overhead}, 
the performance of all baseline evaluations exhibits zero standard deviation across runs due to deterministic representations from OpenCLIP on the split. 
From the table, we observe that using textual modifications to retrieve target images yields slightly better performance than using the reference image alone. The performance of the combined query falls between these two approaches.
Moreover, representation extraction is computationally efficient for all baseline evaluations. Since none of these processes employ generative models, memory consumption and generative inference time are not reported.
For \pefuse (T$\rightarrow$I), 
we observe that \texttt{Qwen2.5-VL-3B-Instruct} has the best performance among three MLLMs, 
while  \texttt{LLaVA-1.5-7B-hf} performs the worst, which stresses the important role of appropriate MLLMs when converting multimodal queries into text. 
Another valuable observation is that \texttt{Qwen2.5-VL-3B-Instruct} surpasses \texttt{Qwen2.5-VL-7B-Instruct} by 0.18\% for average mAP 
while being slower by 0.28 seconds per sample at inference time.
On the other hand, we utilize \texttt{Qwen2.5-VL-7B-Instruct} for \pefuse (I$\rightarrow$I), for consistency. 
In \autoref{tab:overhead}, \texttt{SDXL-Turbo} is more than $3\times$ faster and outperforms \texttt{SDXL-InstructPix2Pix} by 2.18\% when using raw modifications.
When employing descriptions generated by \texttt{Qwen2.5-VL-7B-Instruct}, a similar phenomenon is observed.
Interestingly, using the MLLM produces performance gains for \texttt{SDXL-InstructPix2Pix} ($\uparrow$ 0.11\%) 
whereas the performance of \texttt{SDXL-Turbo} drops by 0.22\% when conditioned on generated target descriptions.
When using \pefuse (T$\rightarrow$I), the retrieval performance consistently surpasses that of all baselines that do not incorporate generative MLLMs. 
Similarly, when using \pefuse (I$\rightarrow$I), all Diffusion-based models improve retrieval performance compared with using the reference image alone for target image retrieval. 
These results collectively demonstrate the effectiveness of our approach in leveraging generative models to enhance retrieval performance.
Overall, the choice of generative model affects downstream performance when using our method; however, the impact is generally marginal and remains consistent across multiple random seeds when a specific model is selected.

\begin{table}[!ht]
    \centering
    \caption{Ablation and compute latency on the CIRCO validation split over three runs. 
    For baseline performance, all retrieval tasks are deterministic, thus no standard deviation is reported; no generative models are involved, therefore no memory consumption is measured. For \pefuse conversion modes, generative
    inference time and the entire pipeline time are reported \textit{per query}, with mean and standard deviation. 
    ``SDXL-Instr.'' denotes \texttt{SDXL-InstructPix2Pix}.
    }
    \resizebox{\linewidth}{!}{
    \begin{tabular}{lcrrrr}
    \toprule
        \textbf{Method}&\textbf{Generative Models}&\textbf{Memory} (MB)&\textbf{Inference Time} (s)& \textbf{Pipeline Time} (s)& \textbf{Avg mAP} (\%)\\
    \midrule
        Modification$\rightarrow$Target& $-$& $-$&$-$&0.48 ($\pm 0.02$)&7.36\\
        Reference$\rightarrow$Target& $-$& $-$&$-$&0.48 ($\pm 0.02$)&3.48\\
        Reference+modification$\rightarrow$Target& $-$& $-$&$-$&0.49 ($\pm 0.02$)&6.61\\
    \midrule
         \multirow{3}*{\pefuse (T$\rightarrow$I)}
         &Qwen2.5-VL-3B-Instruct&7162&0.69 ($\pm$ 0.03)&1.70 ($\pm$ 0.11)&25.15 ($\pm$ 0.80)\\
         &Qwen2.5-VL-7B-Instruct&15818&0.41 ($\pm$ 0.03)&1.70 ($\pm$ 0.14)&24.97 ($\pm$ 0.37)  \\
         &LLaVA-1.5-7B-hf&13472&0.70 ($\pm$ 0.04)&3.01 ($\pm$ 0.10)&17.11 ($\pm$ 1.09)\\
    \midrule
         \multirow{4}*{\pefuse (I$\rightarrow$I)}
         &SDXL-Instr.&6725&3.95 ($\pm$ 0.02)&6.24 ($\pm$ 0.05)&5.22 ($\pm$ 0.08)\\
         &SDXL-Turbo&6725&1.25 ($\pm$ 0.02)&3.07 ($\pm$ 0.27)&7.40 ($\pm$ 0.56)\\
         &Qwen2.5-VL-7B-Instruct + SDXL-Instr.&15818 + 6725&4.40 ($\pm$ 0.03)&6.33 ($\pm$ 0.23)&5.33 ($\pm$ 0.70) \\
         &Qwen2.5-VL-7B-Instruct + SDXL-Turbo&15818 + 6725&1.69 ($\pm$ 0.05)&3.49 ($\pm$ 0.33)&7.18 ($\pm$ 0.41)\\
    \bottomrule
    \end{tabular}
    }
    \label{tab:overhead}
\end{table}
\section{Model Deployment}
\label{sec:deployment}

Our proposed framework leverages MLLMs and Diffusion models, 
making efficient real-time deployment critical for practical applications.
To ensure low-latency retrieval, we preprocess the entire candidate image pool 
by indexing their visual representations in a vector database using the \texttt{FAISS}~\citep{faiss} library. 
MLLMs or Diffusion models can be used offline to generate new data based on multimodal queries, 
and then store the query representation using the same strategy. 
In this way, decoupling the generative process from the retrieval step can balance performance and efficiency, 
enabling scalable real-time operation.

\section{Limitations}
\label{sec:limitations}

While our method demonstrates promising performance via multiple conversion strategies for ZS-CIR tasks,
and is highly compatible with comparable techniques like LDRE and ImageScope,
it is subject to several challenges inherent to its design, 
which also present opportunities for future research.
First, the integration of Diffusion models and MLLMs for the pseudo-fusion of modalities 
can lead to concept drift issues, which often propagate errors through the pipeline. 
Consequently, chaining multiple models can be error-prone 
and presents a significant challenge for optimizing CIR performance via the entire pipeline.
Second, our framework considers non-preprocessed user-provided text modifiers as inputs to MLLMs. 
We posit that performance could be enhanced through advanced text paraphrasing, structured formatting, or more sophisticated prompt engineering strategies.
However, such techniques must be carefully designed to mitigate the inherent risk of LLM hallucinations.
Third, Diffusion models and MLLMs are highly sensitive to their hyperparameters;
nevertheless, tuning these hyperparameters is a labor-intensive process 
that may not generalize across diverse datasets.
Ultimately, generating image captions via MLLMs or new images via Diffusion models can be computationally expensive at web scale,
which is a common limitation for LLM/MLLM-based methods~\citep{cirevl, ldre, imagescope, weimocir}.

Based on these limitations, we identify several promising directions for future work. 
Efforts could focus on developing robust integration techniques 
to minimize error propagation in multi-model pipelines. 
Furthermore, considering more complex scenarios remains a compelling long-term goal. 
For example, compositions involving multiple input images, longer text narratives, or additional modalities such as video, are worth investigating.
Another promising avenue is to explore iterative, multi-round fusion of images and text 
to generate progressively more refined and more accurate descriptions.
Finally, such conversion paradigms inherently introduce some latency. 
Therefore, to deploy such interpretable CIR systems at scale, 
more sophisticated and lightweight models should be developed.

\section{Conclusion}
\label{sec:conclusion}

This work represents the first systematic investigation and benchmarking of training-free pseudo-fusion 
for both uni-directional and bi-directional conversion within ZS-CIR. 
We empirically quantified the relationship 
between CIR performance and key hyperparameters of modern generative models, 
and conducted an ablation and latency analysis when using various MLLMs and Diffusion models.
Our results demonstrate that CIR tasks can be effectively reformulated 
by converting heterogeneous modalities into a single, unified modality. 
This approach enables the use of standard single-query retrieval systems, 
either intra-modal or cross-modal, leveraging existing high-performance pretrained models in a plug-and-play manner, without having to train new modules.
Furthermore, our analysis establishes that reformulating the CIR task as a text-to-image retrieval task is a more effective strategy compared with other conversion modes, emphasizing the importance of choosing appropriate MLLMs and Diffusion models when converting multimodal queries. 
The competitive performance of generative models in this pseudo-fusion role 
underscores their potential as a powerful tool for modality unification 
and points to a promising future for generative, model-based fusion methods in multimodal machine learning.
\section*{Acknowledgments}

Research supported by the Luxembourg National Research Fund (FNR)
through the PRIDE Doctoral Training Unit ``Deep Data Science of Digital History'' (PRIDE21/16758026/D4H)
and the European Innovation Council through the Pathfinder program (SYMBIOTIK project, grant 101071147).

The experiments presented in this paper were carried out using 
the HPC facilities of Luxembourg’s national supercomputer MeluXina. 
The authors gratefully acknowledge the LuxProvide team for their expert support.

\bibliographystyle{tmlr}
\bibliography{ref}

\appendix
\section{Datasets}
\label{app:dataset}

\autoref{tab:dataset} shows the details of each dataset we used in our experiments. 
Due to broken links in the Fashion-IQ dataset, some reference images from the original dataset are missing:
98 images for shirt, 158 images for dress, and 97 for toptee.
CIRR, CIRCO, and GeneCIS datasets have all images available.

\begin{table}[!ht]
    \centering
    \caption{Public benchmarks used in our experiments. We use the validation split for Fashion-IQ and test splits for CIRCO, CIRR, and GeneCIS.}
    \begin{tabular}{lrr}
        \toprule
         \bf Dataset& \bf \# of Queries & \bf \# of Candidates  \\
        \midrule
         Fashion-IQ (Shirt) &1940&6181 \\
         Fashion-IQ (Dress) &1859&3648 \\
         Fashion-IQ (Toptee) &1867&5261 \\
         CIRCO &800&123403 \\
         CIRR&4148&2315 \\
         GeneCIS (Focus Attribute)&2000&20000 \\
         GeneCIS (Change Attribute)&2112&31680 \\
         GeneCIS (Focus Object)&1960&29400 \\
         GeneCIS (Change Object)&1960&29400 \\
        \bottomrule
    \end{tabular}
    \label{tab:dataset}
\end{table}
\section{Training-based Methods}
\label{app:traning_methods}

To gain further insights,
we provide the baseline performance of training-based methods on all datasets in \autoref{tab:fashioniq_perf_training}, \autoref{tab:circo_perf_training}, \autoref{tab:cirr_perf_training}, and \autoref{tab:genecis_training}, respectively.
Nonetheless, we mainly focus on training-free methods in this paper.

\begin{table}[!ht]
    \centering
    \caption{Performance (\%) of training-based methods on \textbf{Fashion-IQ} validation split. \textbf{Best} and \underline{second} best highlighted. $^{*}$: reproduced results; $^{\dag}$: results from original papers.
    } 
    \begin{tabular}[width=\linewidth]{llrr rr rr rr}
        \toprule
         \multirow{2}*{\bf Method}&\multirow{2}*{\bf Retrieval Model}&\multicolumn{2}{c}{\bf Shirt}&\multicolumn{2}{c}{\bf Dress}&\multicolumn{2}{c}{\bf Toptee}&\multicolumn{2}{c}{\bf Average}  \\
         \cmidrule(lr){3-4}\cmidrule(lr){5-6}\cmidrule(lr){7-8}\cmidrule(lr){9-10}
         &&R@10&R@50&R@10&R@50&R@10&R@50&R@10&R@50 \\
         
        \midrule
            Pic2Word$^{\dag}$&CLIP (ViT-L/14)&26.20&43.60&20.00&40.20&27.90&47.40&24.70&43.70 \\
            SEARLE-OTI$^{*}$&CLIP (ViT-B/32)&24.43&41.39&19.85&40.72&24.85&45.47&23.05&42.53\\
            SEARLE$^{*}$&CLIP (ViT-B/32)&24.85&41.60&19.37&39.21&25.12&46.22&23.11&42.34\\
            CoLLM$^{\dag}$&CLIP (ViT-B/32)&24.90&45.10&22.90&43.80&26.40&46.80&24.80&45.20 \\
            LinCIR$^{*}$&CLIP (ViT-L/14)&29.69&48.14&22.32&45.13&30.85&52.01&27.62&48.43 \\
            SEARLE+CIG-XL turbo$^{\dag}$&CLIP (ViT-B/32)&24.73&41.46&18.94&39.66&25.50&46.66&23.06 &42.59\\
            HyCIR$^{\dag}$&CLIP (ViT-L/14)&27.62&44.94&19.98&40.80&28.14&47.67&25.25&44.47 \\
        \bottomrule
    \end{tabular}
    \label{tab:fashioniq_perf_training}
\end{table}

\begin{table}[!ht]
    \centering
    \caption{Performance (\%) of training-based methods on \textbf{CIRR} test split. 
    $^{*}$: reproduced results; $^{\dag}$: results from original papers; $-$: results not available.
    }
    \begin{tabular}{ll rrrrr rrr}
        \toprule
         \multirow{2}*{\bf Method}&\multirow{2}*{\bf Retrieval Model}&\multicolumn{5}{c}{\bf Recall}&\multicolumn{3}{c}{\bf Recall$_{\text{subset}}$}\\
         \cmidrule(lr){3-7}\cmidrule(lr){8-10}
         &&@1&@2&@5&@10&@50&@1&@2&@3 \\
        \midrule
         Pic2Word$^{\dag}$&CLIP (ViT-L/14)&23.90&\textemdash&51.70&65.30&87.80&53.76&74.46&87.08 \\
         SEARLE-OTI$^{*}$&CLIP (ViT-B/32)&23.18&34.72&52.31&66.00&89.21&52.02&74.43&86.75 \\
         SEARLE$^{*}$&CLIP (ViT-B/32)&23.33&34.89&52.89&66.99&89.81&53.90&76.19&87.76 \\
         CoLLM$^{\dag}$&CLIP (ViT-B/32)&28.60&\textemdash&\textemdash&71.80&92.70&\textemdash&\textemdash&\textemdash \\
         LinCIR$^{*}$&CLIP (ViT-L/14)&25.04&36.22&53.78&67.18&88.75&56.53&76.82&88.70 \\
         SEARLE+CIG-XL turbo$^{\dag}$&CLIP (ViT-B/32)&25.54&\textemdash&55.01&68.24&90.72&57.52&78.36&89.35\\
         HyCIR$^{\dag}$&CLIP (ViT-L/14)&25.08&\textemdash&53.49&67.03&89.85&53.83&75.06&87.18 \\
        \bottomrule
    \end{tabular}
    \label{tab:cirr_perf_training}
\end{table}

\begin{table}[!ht]
    \centering    
    \caption{Performance (\%)  of training-based methods on \textbf{CIRCO} test split.
    $^{*}$: reproduced results; $^{\dag}$: results from original papers; $-$: results not available.
    }
    \begin{tabular}{llrrrr}
        \toprule
         \bf Method&\bf Retrieval Model&\bf mAP@5&\bf mAP@10&\bf mAP@25&\bf mAP@50 \\
        \midrule
        Pic2Word$^{\ddagger}$&CLIP (ViT-L/14)&8.72&9.51&10.64&11.29 \\
        SEARLE-OTI$^{*}$&CLIP (ViT-B/32)&7.29&7.99&9.21&9.85 \\
        SEARLE$^{*}$&CLIP (ViT-B/32) &9.38&9.95&11.13&11.85 \\
        CoLLM$^{\dag}$&CLIP (ViT-B/32)&12.90&13.20&\textemdash&15.00 \\
        LinCIR$^{*}$ &CLIP (ViT-L/14)&12.33&13.13&14.56&15.46 \\
        SEARLE+CIG-XL turbo$^{\dag}$&CLIP (ViT-B/32)&10.45&11.02&12.34&13.00\\
        HyCIR$^{\dag}$&CLIP (ViT-L/14)&14.12&15.02&16.72&17.56 \\
        \bottomrule
    \end{tabular}
    \label{tab:circo_perf_training}
\end{table}

\begin{table}[!ht]
    \centering
    \caption{Performance (\%)  of training-based methods on \textbf{GeneCIS} test split.}
    \resizebox{\linewidth}{!}{
    \begin{tabular}{ll rrr rrr rrr rrr r}
        \toprule
         \multirow{2}*{\textbf{Method}}&\multirow{2}*{\textbf{Retrieval Model}}&\multicolumn{3}{c}{\textbf{Focus Attribute}}&\multicolumn{3}{c}{\textbf{Change Attribute}}&\multicolumn{3}{c}{\textbf{Focus Object}}&\multicolumn{3}{c}{\textbf{Change Object}}&\textbf{Average}  \\
         \cmidrule(lr){3-5}\cmidrule(lr){6-8}\cmidrule(lr){9-11}\cmidrule(lr){12-14}
         &&R@1&R@2&R@3&R@1&R@2&R@3&R@1&R@2&R@3&R@1&R@2&R@3&R@1\\
        \midrule
        Pic2Word$^\dag$&CLIP (ViT-L/14)&15.65&28.16&38.65&13.87&24.67&33.05&8.42&18.01&25.77&6.68&15.05&24.03&11.16\\
        SEARLE$^\dag$&CLIP (ViT-B/32)&18.90&30.60&41.20&13.00&23.80&33.70&12.20&23.00&33.30&13.60&23.80&33.30&14.40\\
        LinCIR$^\dag$&CLIP (ViT-L/14)&16.90&29.95&41.45&16.19&27.98&36.84&8.27&17.40&26.22&7.40&15.71&25.00&12.19\\
        LinCIR+CIG-XL turbo$^\dag$&CLIP (ViT-L/14)&16.80&29.70&40.90&15.91&28.88&37.45&8.37&17.35&25.10&7.86&15.46&24.29&12.24\\
        \bottomrule
    \end{tabular}
    }
    \label{tab:genecis_training}
\end{table}
\section{Scaling Laws on Fashion-IQ Categories, CIRR Subsets, and GeneCIS Categories}
\label{app:scaling_law_ext}

We report results of scaling laws for each category of Fashion-IQ dataset and the subsets of CIRR in \autoref{tab:scaling_law_fashioniq_cat} and those of GeneCIS in \autoref{tab:scaling_law_genecis_cat}.

\begin{table}[!ht]
    \centering
    \caption{Scaling laws on each category of \textbf{Fashion-IQ} and \textbf{CIRR} subsets using OpenCLIP for the text-to-image retrieval task.}
    \begin{tabular}{rrrrrrrrrr}
    \toprule
    \multirow{2}*{\bf Backbone}&\multicolumn{2}{c}{\bf Shirt}&\multicolumn{2}{c}{\bf Dress}&\multicolumn{2}{c}{\bf Toptee}&\multicolumn{3}{c}{\bf CIRR}\\
    \cmidrule(lr){2-3}\cmidrule(lr){4-5}\cmidrule(lr){6-7}\cmidrule(lr){8-10}
    &R@10&R@50&R@10&R@50&R@10&R@50&R$_{\text{subset}}$@1&R$_{\text{subset}}$@2&R$_{\text{subset}}$@3\\
    \midrule
     ViT-L/14&29.54&47.01&25.23&43.73&33.69&54.37&73.49&88.82&95.08\\
    ViT-H/14&30.41&47.22&26.90&47.71&33.90&55.33&74.41&89.23&95.33\\
    ViT-g/14&30.52&48.87&25.12&46.05&34.82&55.44&74.15&89.57&95.45\\
    ViT-bigG/14&31.39&47.32&25.12&45.56&34.82&55.60&75.90&89.37&95.59\\
    \bottomrule
    \end{tabular}
    \label{tab:scaling_law_fashioniq_cat}
\end{table}

\begin{table}[!ht]
    \centering
    \caption{Scaling laws on each category of \textbf{GeneCIS} using OpenCLIP for text-to-image task.}
    \begin{tabular}{rrrrrrrrrrrrr}
    \toprule
    \multirow{2}*{\bf Backbone}&\multicolumn{3}{c}{\bf Focus Attribute}&\multicolumn{3}{c}{\bf Change Attribute}&\multicolumn{3}{c}{\bf Focus Object}&\multicolumn{3}{c}{\bf Change Object}\\
    \cmidrule(lr){2-4}\cmidrule(lr){5-7}\cmidrule(lr){8-10}\cmidrule(lr){11-13}
    &R@1&R@2&R@3&R@1&R@2&R@3&R@1&R@2&R@3&R@1&R@2&R@3\\
    \midrule
     ViT-L/14&17.85&30.30&41.90&14.25&27.27&37.12&16.63&27.50&38.21&18.67&31.02&40.61\\
    ViT-H/14&19.10&31.35&43.50&15.58&27.18&37.97&17.50&28.16&37.09&17.81&29.59&39.54\\
    ViT-g/14&19.25&32.05&42.10&15.77&27.94&37.83&17.35&27.50&37.24&17.19&29.69&39.39\\
    ViT-bigG/14&19.00&31.15&42.95&16.57&28.79&39.44&17.55&27.91&37.45&18.83&30.77&40.87\\
    \bottomrule
    \end{tabular}
    \label{tab:scaling_law_genecis_cat}
\end{table}
\section{Prompts}\label{app:prompts}

We show the prompts used for MLLMs when generating composed descriptions based on reference images and text modifications.
When using the prompts, 
images are converted to base64 format and then inserted into the prompts.
We point out that MLLMs can also be employed to generate multiple descriptions from different aspects or views per query like in LDRE~\citep{ldre}, 
which further enhances CIR performance at the cost of extra computational overhead.

\begin{figure}[!ht]
    \centering
\cbox[Fashion-IQ]{
You are an expert at visual perception and imagination of fashion items. 
Given a reference image of fashion items and modification instructions, mentally apply the changes and produce an accurate and complete natural-language description of the resulting fashion items. 
The modifications may describe direct attributes (e.g., “solid white with buttons”), comparisons (e.g., “longer sleeves,” “lighter in color”), combined attributes (e.g., “black with a red cherry pattern and deep V neckline”), or negations (e.g., “no lace design”).
Image: \textcolor{ForestGreen}{\texttt{base64\_image}}.
Here are the modification instructions: \textcolor{ForestGreen}{\texttt{caption}}.
Focus on the fashion item and its attributes such as type, color, pattern, material, shape, fit, and style details.
Ignore people and background from the image.
Avoid imaginary things. 
Be specific and objective so that I can find targeting images based on your description solely without knowing the reference image or modification instructions.
Do not use vague comparative terms like 'same/different/smaller/larger/shorter/longer/unchanged', etc. Instead, you should specify these differences clearly, like: another color instead of red (if no specific targeting color is mentioned), and a clear sky (if mentioned) instead of unchanged sky, etc.
Now, describe how the final fashion item looks after applying the modifications. 
Write in 1 to 3 coherent sentences.
}
\end{figure}

\begin{figure}[!ht]
\centering
\cbox[CIRR]{
You are an expert at visual imagination of real-world scenes. 
Given a reference image and modification instructions, mentally apply the modifications to the reference image and describe the resulting image in clear, complete English. 
Apply the modifications exactly as described, and ensure the final description reflects the scene after the changes.
The modifications may include:
1. Cardinality: adjusting the number of objects (e.g., “only one bird remains”).
2. Addition: adding new objects or attributes (e.g., “add a red chair in the corner”).
3. Negation: removing elements (e.g., “remove the table”).
4. Direct Addressing: ensuring specific mentioned objects are clearly included.
5. Compare \& Change: replacing one attribute with another (e.g., “same sofa but in leather”).
6. Comparative Statement: relative size, quantity, or intensity changes (e.g., “a larger group of people”).
7. Conjunction Statements: multiple modifications combined (e.g., “remove the tree and add two benches”).
8. Spatial Relations \& Background: modifying positions, layout, or setting (e.g., “change the background to a beach”).
9. Viewpoint: adjusting perspective or framing (e.g., “zoom out to show the whole scene”).
Image: \textcolor{ForestGreen}{\texttt{base64\_image}}.
Here are the modification instructions: \textcolor{ForestGreen}{\texttt{caption}}.
Focus on the elements (like objects, people, and animals), their attributes (like color, size, shape, and quantity), spatial relations, and background.
Avoid imaginary details and unnecessary repetitions.
Be specific and objective so that I can find the target image from an image gallery based on your description solely without knowing the reference image or modification instructions.
Do not use vague comparative terms like 'same/different/smaller/larger/shorter/longer/unchanged', etc. Instead, you should specify these differences clearly, like: another color instead of red (if no specific targeting color is mentioned), and a clear sky (if mentioned) instead of unchanged sky, etc.
Write in 1 to 3 coherent sentences in English.
Now, describe how the final image looks after applying the modifications. 
}
\end{figure}

\begin{figure}[!ht]
    \centering
\cbox[CIRCO]{
You are an expert at visual imagination of real-world scenes. 
Given a reference image and modification instructions, mentally apply the modifications and produce an accurate, detailed description of the resulting scene. 
Apply the modifications exactly as described, and describe the final scene after the changes.
The modifications may involve:
1. Cardinality: adjusting the number of objects (e.g., “has two boxes”).
2. Addition: introducing new objects or attributes (e.g., “a child under the umbrella”).
3. Negation: removing elements (e.g., “shows no bike”).
4. Direct Addressing: ensuring a specific object is present (e.g., “next to a window”).
5. Compare \& Change: altering attributes (e.g., “different color,” “surrounded by flowers”).
6. Comparative Statements: relative size, number, or intensity (e.g., “more stickers,” “larger crowd”).
7. Conjunction Statements: multiple edits at once (e.g., “surrounded by snow and trees are more bare”).
8. Spatial Relations \& Background: positioning or environment changes (e.g., “skyscrapers in the background”).
9. Viewpoint: changes in perspective or framing (e.g., “shot from above”).
Image: \textcolor{ForestGreen}{\texttt{base64\_image}}.
Here are the modification instructions: \textcolor{ForestGreen}{\texttt{caption}}.
Focus on the objects, people, animals, attributes (color, size, shape, quantity), spatial relations, and background context. 
Be specific and objective. 
Avoid imaginary details not supported by the reference image or the modification. 
Do not use vague comparative terms like 'same/different/smaller/larger/shorter/longer/unchanged', etc. Instead, you should specify these differences clearly, like: another color instead of red (if no specific targeting color is mentioned), and a clear sky (if mentioned) instead of unchanged sky, etc.
Write 1 to 3 complete and coherent sentences so that I can find targeting images based on your description solely without knowing the reference image or modification instructions.
Now, describe how the final image looks after applying these modifications.
}
\end{figure}

\begin{figure}[!ht]
    \centering
\cbox[GeneCIS]{
You are an expert at visual imagination of real-world scenes. 
Given a reference image and modification instructions, you should distinguish the objects and their attributes from the reference image, and then mentally apply the modifications to the reference image, and describe the objects and their attributes in the final image after the changes. 
Image: \textcolor{ForestGreen}{\texttt{base64\_image}}.
Here are the modification instructions: \textcolor{ForestGreen}{\texttt{caption}}.
Avoid imaginary details not supported by the reference image or the modification. 
Write 1 to 3 complete and coherent sentences so that I can find targeting images based on your description solely without knowing the reference image or modification instructions.
Now, describe the objects and their attributes after applying the modification.
}
\end{figure}
\section{Qualitative Results Using MLLM for Diffusion Models}
\label{app:qualitative}

\begin{figure}[!ht]
    \centering
    \def\w{0.44\linewidth} 
  \begin{subfigure}{\w}
    \centering
    \includegraphics[width=\linewidth]{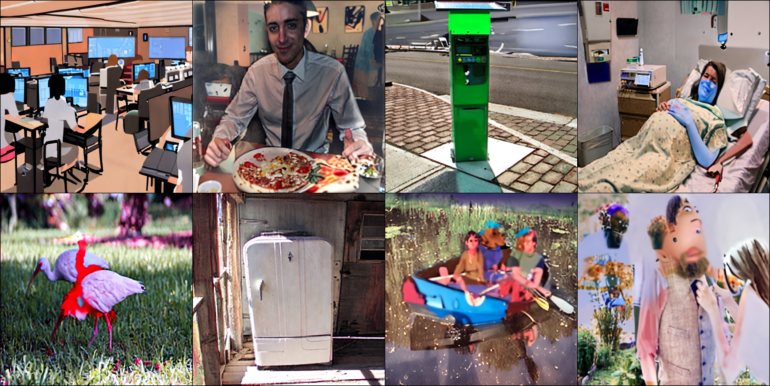}
    \label{fig:circo_no_7_ins}
  \end{subfigure}
  \hfill
  \begin{subfigure}{\w}
    \centering
    \includegraphics[width=\linewidth]{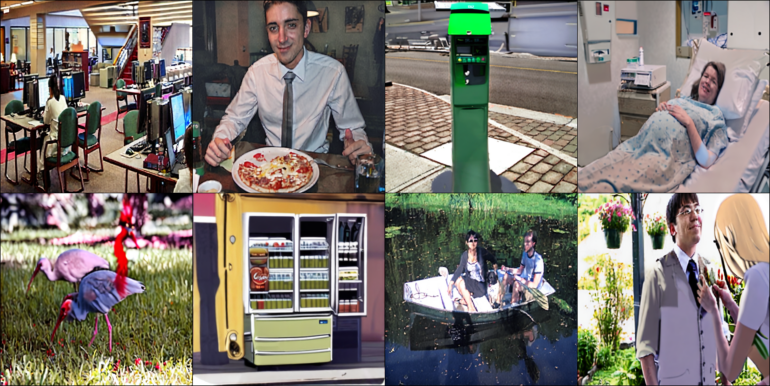}
    \label{fig:circo_yes_7_ins}
  \end{subfigure}
  \begin{subfigure}{\w}
    \centering
    \includegraphics[width=\linewidth]{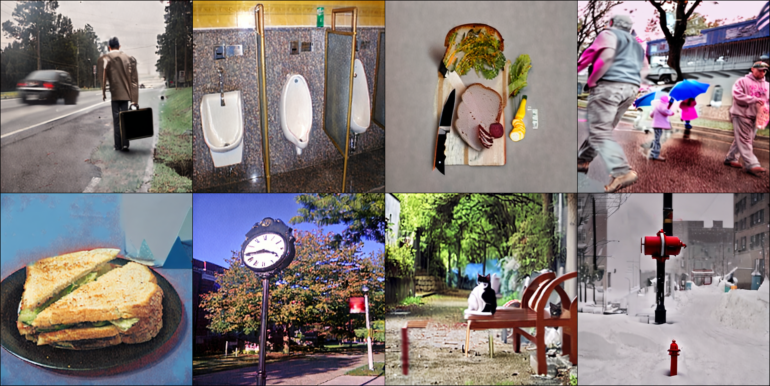}
    \label{fig:circo_no_17_ins}
  \end{subfigure}
  \hfill
  \begin{subfigure}{\w}
    \centering
    \includegraphics[width=\linewidth]{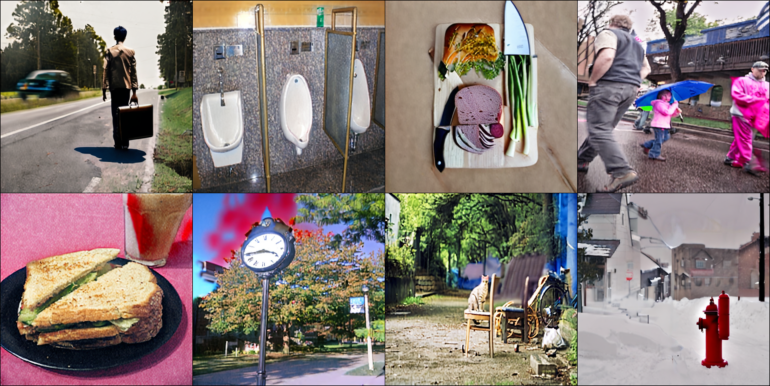}
    \label{fig:circo_yes_17_ins}
  \end{subfigure}
  \begin{subfigure}{\w}
    \centering
    \includegraphics[width=\linewidth]{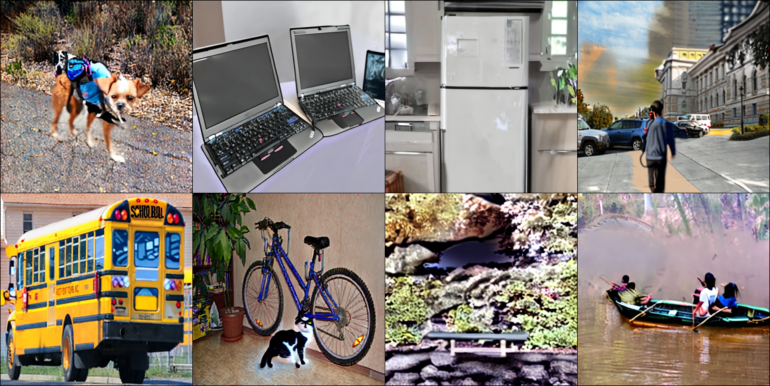}
    \caption{With raw captions.}
    \label{fig:circo_no_18_ins}
  \end{subfigure}
  \hfill
  \begin{subfigure}{\w}
    \centering
    \includegraphics[width=\linewidth]{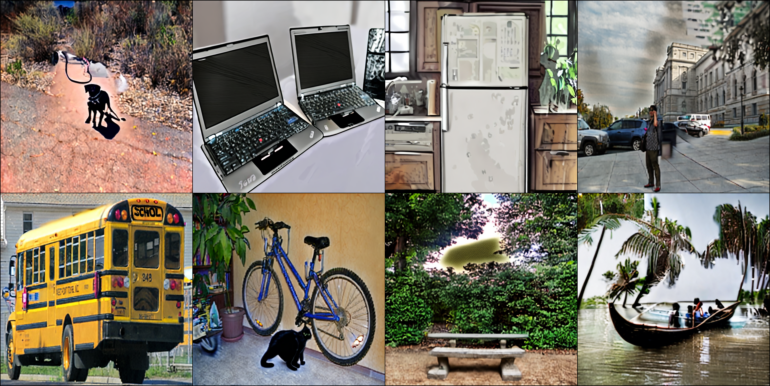}
    \caption{With MLLM-generated descriptions.}
    \label{fig:circo_yes_18_ins}
  \end{subfigure}
\caption{Qualitative results of \textbf{SDXL-InstructPix2Pix}-generated images with Qwen2.5-VL-7B-generated descriptions and with raw captions from the CIRCO validation split. We use 3.0 for image guidance scale, 7.5 for guidance scale, and 30 denoising steps.}
\label{fig:qunlitative_circo_instruct}
\end{figure}

\begin{figure}[!ht]
    \centering
    \def\w{0.44\linewidth} 
  \begin{subfigure}{\w}
    \centering
    \includegraphics[width=\linewidth]{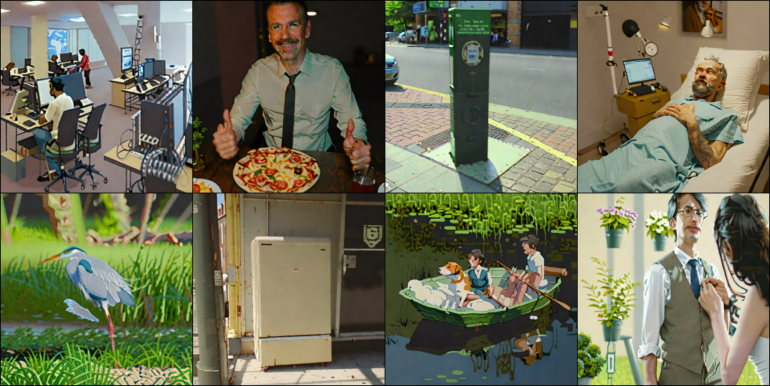}
    \label{fig:circo_no_7_tur}
  \end{subfigure}
  \hfill
  \begin{subfigure}{\w}
    \centering
    \includegraphics[width=\linewidth]{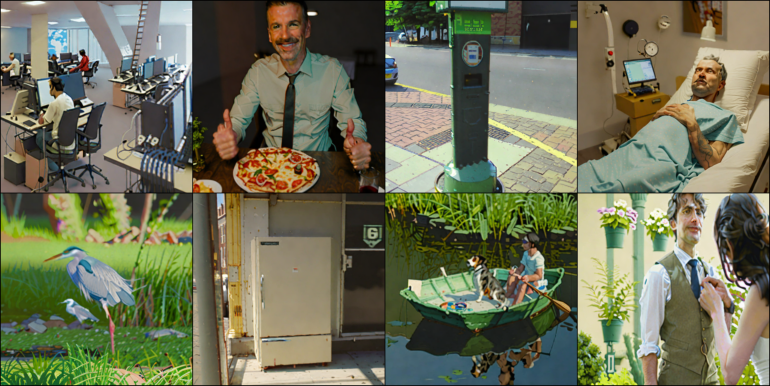}
    \label{fig:circo_yes_7_tur}
  \end{subfigure}
  \begin{subfigure}{\w}
    \centering
    \includegraphics[width=\linewidth]{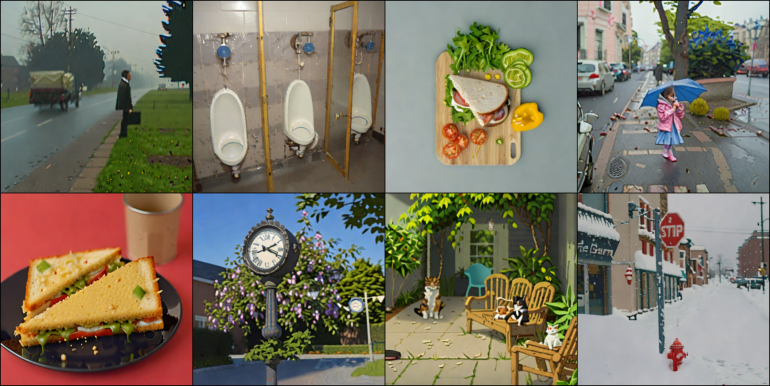}
    \label{fig:circo_no_17_tur}
  \end{subfigure}
  \hfill
  \begin{subfigure}{\w}
    \centering
    \includegraphics[width=\linewidth]{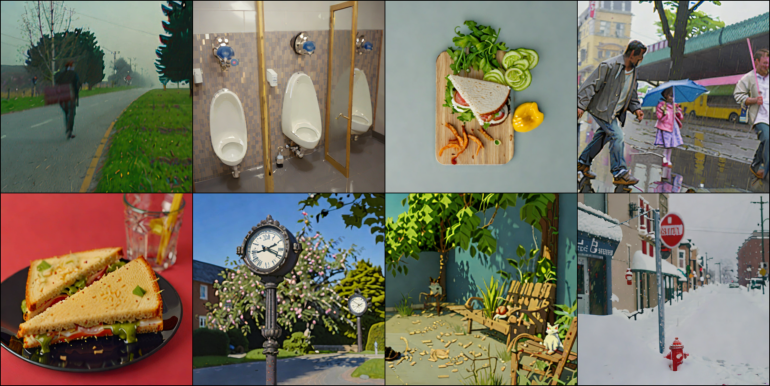}
    \label{fig:circo_yes_17_tur}
  \end{subfigure}
  \begin{subfigure}{\w}
    \centering
    \includegraphics[width=\linewidth]{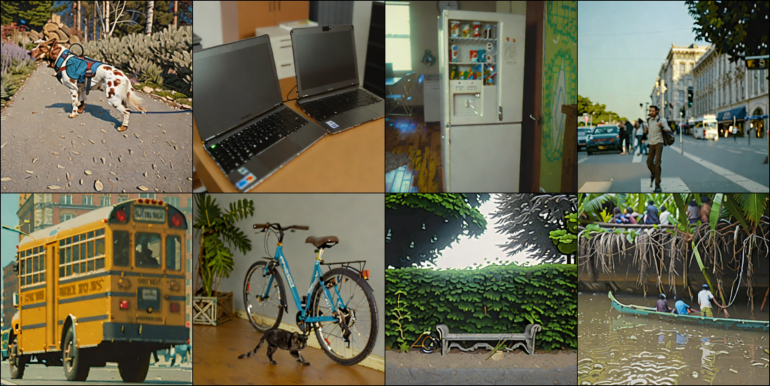}
    \caption{With raw captions.}
    \label{fig:circo_no_18_tur}
  \end{subfigure}
  \hfill
  \begin{subfigure}{\w}
    \centering
    \includegraphics[width=\linewidth]{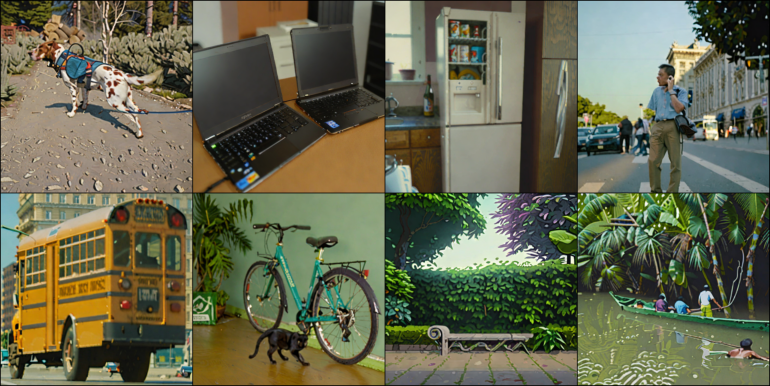}
    \caption{With MLLM-generated descriptions.}
    \label{fig:circo_yes_18_tur}
  \end{subfigure}
\caption{Qualitative results of \textbf{SDXL-Turbo}-generated images with Qwen2.5-VL-7B-generated descriptions and with raw captions from the CIRCO validation split. We use 0.5 for strength, 0.0 for guidance scale, and 30 denoising steps.}
\label{fig:qunlitative_circo_turbo}
\end{figure}

We show the superiority of generated images using MLLM-generated text rather than raw captions for CIRCO validation split in \autoref{fig:qunlitative_circo_instruct}
and \autoref{fig:qunlitative_circo_turbo}.
It can be observed that using raw captions to generate images produces more noticeable artifacts for both Diffusion models, 
and SDXL-Turbo-generated images have more artifacts than those of SDXL-InstructPix2Pix.

\section{Failure Cases}\label{sec:failure}

\begin{figure}[!ht]
    \centering
    \includegraphics[width=.8\linewidth,keepaspectratio=true]{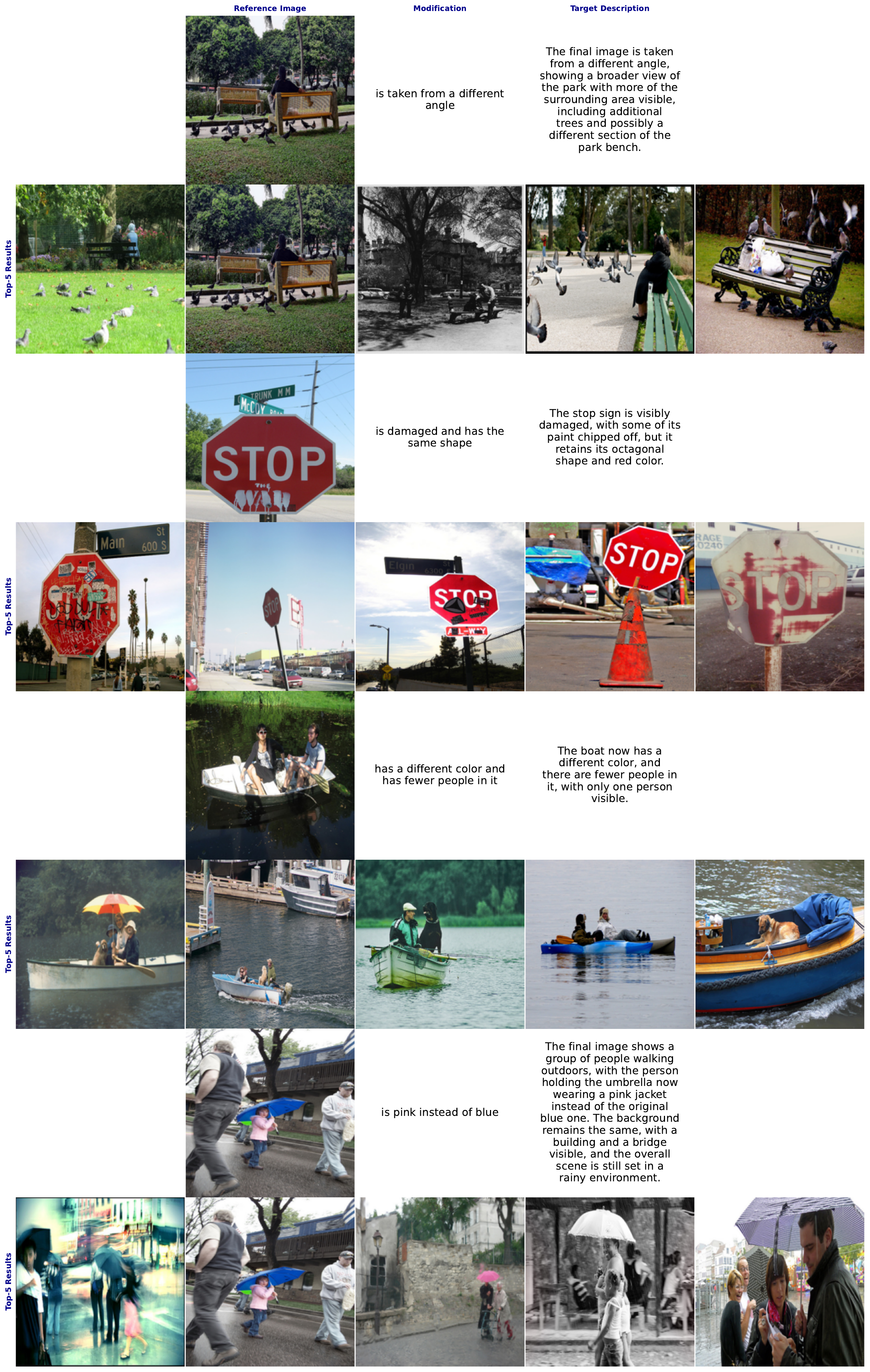}
    \caption{Failure cases of \pefuse (T$\rightarrow$I). None of the top-five results matches the corresponding query.}
    \label{fig:failure_txt2img}
\end{figure}

\begin{figure}[!ht]
    \centering
    \includegraphics[width=0.8\linewidth, keepaspectratio=true]{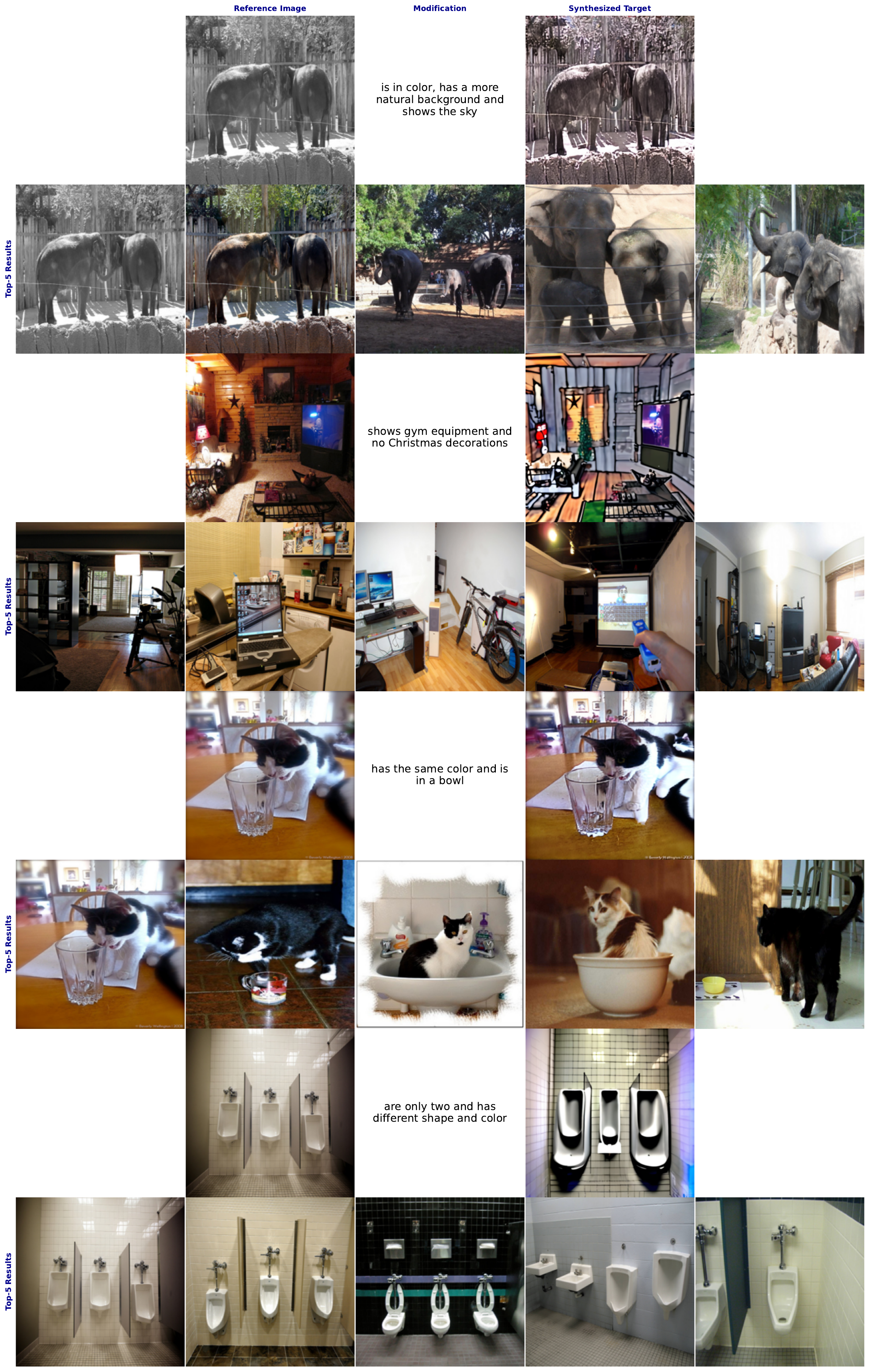}
    \caption{
    Failure cases of \pefuse (I$\rightarrow$I). None of the top-five results matches the corresponding query.
    }
    \label{fig:failure_img2img}
\end{figure}

By way of example, we show the failure cases when using OpenCLIP to retrieve top-5 images from the CIRCO candidate pool of 123k images in \autoref{fig:failure_txt2img} when performing \pefuse (T$\rightarrow$I),
and \autoref{fig:failure_img2img} when performing \pefuse (I$\rightarrow$I).
In \autoref{fig:failure_txt2img}, Qwen-generated composed descriptions are used to retrieve target images. We observe that the generated queries, including queries 1, 2, and 3, capture the semantics of the desired target images, but might be more detailed than expected, causing mismatches with ground-truth images. For query 4, the generated textual query mistakenly identifies the objects to which the modification applies, leading to wrong retrieval results.
In \autoref{fig:failure_img2img}, all the generated images via \texttt{SDXL-InstructPix2Pix} do not fully match the modifications desired for target images, and most of the generated images are very similar to the original reference images, thereby degrading the retrieval performance.

\end{document}